\documentclass[10pt,twocolumn,letterpaper]{article}

\usepackage{cvpr}      

\usepackage{graphicx}
\usepackage{amsmath}
\usepackage{amssymb}
\usepackage{booktabs}

\usepackage{multirow}
\usepackage{wrapfig}
\usepackage{adjustbox}
\usepackage{booktabs}
\usepackage{pifont}
\usepackage{subcaption}

\newcommand{\subtabletopmod}[5]{\parbox[b][#2]{#1}{\footnotesize\centering\vspace{#5}\scalebox{#3}{#4}\vfill}}

\newcommand{\cmark}{\ding{51}}

\newcommand{\bfit}[1]{{\bf{\emph{#1}}}}

\newcommand{\imgnetfid}{12.12}
\newcommand{\imgnetis}{61.17}

\usepackage[pagebackref,breaklinks,colorlinks]{hyperref}

\usepackage[capitalize]{cleveref}
\crefname{section}{Sec.}{Secs.}
\Crefname{section}{Section}{Sections}
\Crefname{table}{Table}{Tables}
\crefname{table}{Tab.}{Tabs.}

\def\cvprPaperID{722} 
\def\confName{CVPR}
\def\confYear{2022}

\begin{document}

\title{The Nuts and Bolts of Adopting Transformer in GANs}
\author{Rui Xu$^{1}$ \hspace{9pt} Xiangyu Xu$^{3}$ \hspace{9pt} Kai Chen$^{2,4}$ \hspace{9pt} Bolei Zhou$^{1}$ \hspace{9pt} Chen Change Loy$^{3}$ \\
	\small{$^{1}$ CUHK-SenseTime Joint Lab, The Chinese University of Hong Kong \hspace{5pt}
      $^{2}$ Shanghai AI Laboratory} \\ 
   \small{$^{3}$ S-Lab, Nanyang Technological University \hspace{5pt}
		$^{4}$ SenseTime Research} \\ 
   {\tt\small xr018@ie.cuhk.edu.hk \hspace{5pt} xiangyu.xu@ntu.edu.sg} \\
   {\tt\small chenkai@sensetime.com \hspace{5pt} bzhou@ie.cuhk.edu.hk \hspace{5pt} ccloy@ntu.edu.sg}
 }

\twocolumn[{
	\renewcommand\twocolumn[1][]{#1}
	\maketitle
	\vspace{-30pt}
	\begin{center}
		\centering
		\includegraphics[width=0.95\linewidth]{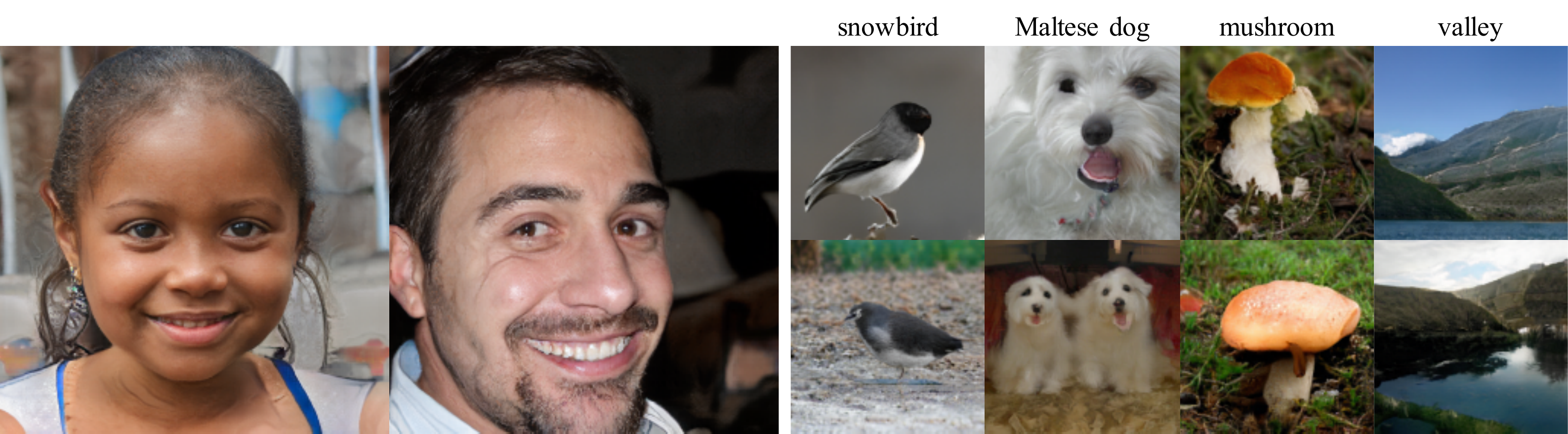}
		\vskip -0.25cm
		\captionof{figure}{Left: uncurated $256\times 256$ results generated for FFHQ with our STrans-G. Right: $128\times 128$ conditional samples from our STrans-G with AdaBN-T injecting \textsc{ImageNet} class information.}
    \vspace{-10pt}
		\label{fig:teaser}
	\end{center}
}]



\begin{abstract}
  \vspace{-8pt}
  Transformer becomes prevalent in computer vision, especially for high-level vision tasks. However, adopting Transformer in the generative adversarial network (GAN) framework is still an open yet challenging problem. In this paper, we conduct a comprehensive empirical study to investigate the properties of Transformer in GAN for high-fidelity image synthesis. 
Our analysis highlights and reaffirms the importance of feature locality in image generation, although the merits of the locality are well known in the classification task.
Perhaps more interestingly, we find the residual connections in self-attention layers harmful for learning Transformer-based discriminators and conditional generators. We carefully examine the influence and propose effective ways to mitigate the negative impacts. 
%
Our study leads to a new alternative design of Transformers in GAN, a convolutional neural network (CNN)-free generator termed as \bfit{STrans-G}, which achieves competitive results in both unconditional and conditional image generations. The Transformer-based discriminator, \bfit{STrans-D}, also significantly reduces its gap against the CNN-based discriminators. 
\end{abstract}

\vspace{-10pt}
\section{Introduction}
\label{sec:intro}

Transformers~\cite{vaswani2017attention} have shown remarkable performance in natural language processing, and lately in computer vision, thanks to the exceptional representation learning capability of self-attention layers. The success has led to further exploration of replacing the commonly-used convolutional neural network (CNN) backbone with Transformers in generative adversarial networks (GAN)~\cite{goodfellow2014generative} for image synthesis. Early attempts~\cite{jiang2021transgan,lee2021vitgan} revealed that it is challenging to design an effective Transformer-based GAN. In particular, a direct application of Transformers previously designed for classification tasks in GAN often leads to inferior synthesis performance to the CNN counterparts. For instance, the GAN model with a ViT backbone~\cite{jiang2021transgan} only achieves a Fr\'{e}chet inception distance (FID)~\cite{heusel2017gans} of 8.92 in $64 \times 64$ \textsc{CelebA} dataset, compared to the FID of 3.16 achieved by the CNN-based StyleGAN2~\cite{karras2020analyzing}.
Besides, the Transformer structure makes GAN training unstable, requiring manual tunings of hyper-parameters.

This work aims to understand and address the potential issues of adopting Transformers in GANs for high-fidelity image generation. We investigate not only unconditional image synthesis but also the less-explored conditional setting. With careful designs, we show that the performance gap can be greatly reduced between the Transformer-based GANs and those based on CNN backbones. 
Our study provides the following `nuts and bolts' of adopting Transformer in GANs:

\noindent
1) \textit{Locality is essential} - Locality of feature extraction has been shown pivotal to efficiency and performance of Transformer in image classification~\cite{li2021localvit,d2021convit}. We find that such a rule-of-thumb is also applicable to GAN-based image generation. In particular, we show that global self-attention operations~\cite{vaswani2017attention,dosovitskiy2020image} as practiced in existing Transformer-based GANs~\cite{jiang2021transgan} are detrimental to the synthesis performance as well as computationally prohibitive for high-fidelity image generation. We examine several alternatives of adding locality to Transformers. Among these methods, Swin layer~\cite{liu2021swin} proves to be the most effective building block to offer the locality inductive bias. It is noteworthy that while all the findings above appear seemingly trivial, no existing studies have provided a similar analysis.

\noindent
2) \textit{Mind the residual connections in the discriminator} - Transformer employs a residual connection around each sub-layer of self-attention and the pointwise fully connected layer. Through a detailed analysis of norm ratios, we find that residual connections tend to dominate the information flow in a Transformer-based discriminator. Consequently, sub-layers that perform self-attention and fully connected operations in the discriminator are inadvertently bypassed, causing inferior quality and slow convergence during training.  
We address this problem by replacing each residual connection with a skip-projection layer, which better retains the information flow in the residual blocks. 

\noindent
3) \textit{Transformer-specific strategy for placing conditional normalization} - We observe that conventional approaches of injecting conditional class information do not work well for Transformer-based conditional GAN. The main culprit lies in the large information flow through the residual connections in the Transformer generator. If the conditional information is injected within the main branch, which is also known as the residual mapping path in \cite{he2016deep}, it will largely be ignored and contribute little to the final outputs. We present a viable way of adopting conditional normalization layers in the trunk, which helps retain conditional information throughout the Transformer generator.

Taking the principles above into consideration, we successfully reduce the gaps between Transformer-based GAN and contemporary CNN-based GANs, which were deemed difficult in previous studies. All the design choices can be implemented easily without elaborative architectural modifications to the commonly-used Transformer models. The resulting model, called \bfit{STransGAN}, achieves a state-of-the-art FID of 2.03 in \textsc{CelebA} dataset and reaches a competitive FID of 4.84 in FFHQ-256 dataset. 
In our attempt of using Transformers for conditional synthesis, the presented model improves the inception score from 10.14 to 11.62 in the challenging CIFAR10 dataset and achieves competitive performance in \textsc{ImageNet1k} (Fig.~\ref{fig:teaser}). It is noteworthy that this study presents the first successful application of Transformer-based GAN under the conditional setting.


\vspace{-2pt}
\section{Related Work}
\vspace{-2pt}

\noindent
\textbf{Vision Transformer.}
Due to its success in NLP, Transformer~\cite{vaswani2017attention} has obtained growing attention in the computer vision community~\cite{lin2021cvpr, carion2020end,xu2021texformer}.
ViT~\cite{dosovitskiy2020image} and subsequent studies~\cite{touvron2021training,zhou2021deepvit,rao2021dynamicvit} successfully transfer the Transformer architecture to the image domain. 
Transformer-based models have been proven effective in multiple high-level tasks, \eg, image detection~\cite{carion2020end,zhu2020deformable} and segmentation~\cite{zheng2021rethinking,strudel2021segmenter}.
However, the standard Transformer with heavy computational costs prevents its application on high-resolution images.
To reduce the computational costs, recent studies propose to use local attention~\cite{liu2021swin, zhang2021aggregating}, which shows superior performance.
%
%
There have been a few recent efforts in image restoration~\cite{chen2021pre,liang2021swinir} and image synthesis~\cite{esser2021taming} with Transformer. Nevertheless, they merely apply the Transformer block as an intermediate component, while our generator is a CNN-free model with only Transformer blocks.

\noindent
\textbf{Generative Adversarial Networks.}
Convolutional GANs~\cite{radford2015unsupervised, arjovsky2017wasserstein} are dominant in the literature.
\cite{karras2017progressive} develop a progressive growing architecture to facilitate higher-resolution synthesis.
The state-of-the-art StyleGAN2~\cite{karras2019style, karras2020analyzing} offer an architecture to disentangle the style information.
In the conditional GANs, the attention mechanism~\cite{zhang2019self,brock2018large,kang2020contragan} has been widely adopted in convolutional generators to better capture the shape priors and long-range correspondence.
However, these works merely regard the self-attention block as an auxiliary module.
In this study, we carefully study the role and impact of self-attention blocks in Transformers when the architecture is used as the generator and discriminator.

\begin{table}[t]
    \centering
    \caption{Comparison between concurrent work and ours. We summarize if these studies investigate the generator (Gen.) or/and discriminator (Disc.) design. Additionally, we show whether unconditional (Uncond.) and conditional (Cond.) synthesis are supported.}
    \vspace{-9pt}
    \begin{adjustbox}{width=0.87\linewidth,center}
    \begin{tabular}{c|c|c|c|c}
    \bottomrule
    \multirow{2}{*}{\textbf{Methods}} & \multicolumn{2}{c|}{Architecture} & \multicolumn{2}{c}{Sampling} \\ \cline{2-5} 
                             & Gen.               & Disc.               & Uncond.        & Cond.        \\ \hline \hline
    TwoTransGANs ~\cite{jiang2021transgan}            &     \cmark       &     \cmark       &     \cmark      &              \\ \hline
    HiT ~\cite{zhao2021improved}                     &     \cmark       &                 &       \cmark    &              \\ \hline
    ViTGANs~\cite{lee2021vitgan}                  &     \cmark       &    \cmark        &     \cmark      &              \\ \hline
    StyleFormer~\cite{park2021styleformer}                  &     \cmark       &          &     \cmark      &              \\ \hline
    Ours                     &  \cmark         &  \cmark         &  \cmark        &    \cmark    \\ \toprule
    \end{tabular}
    \end{adjustbox}
    \label{tab:related}
    \vspace{-18pt}
\end{table}

\noindent
\textbf{Transformer in GANs.}
The most related works to ours are several pilot studies~\cite{jiang2021transgan,zhao2021improved,lee2021vitgan,park2021styleformer}, which also attempt to devise Transformer-based GANs.
Apart from the difference in adopting Swin Transformer, we present a detailed comparison between these concurrent works and ours in Tab.~\ref{tab:related}.
In this work, we conduct a more comprehensive study containing the conditional generation setting.
Besides, we investigate the phenomenon of imbalanced feature flow in Transformer blocks and its impacts to the discriminators and conditional generators.

\begin{figure*}[t]
    
    \begin{center}
    \includegraphics[width=\linewidth]{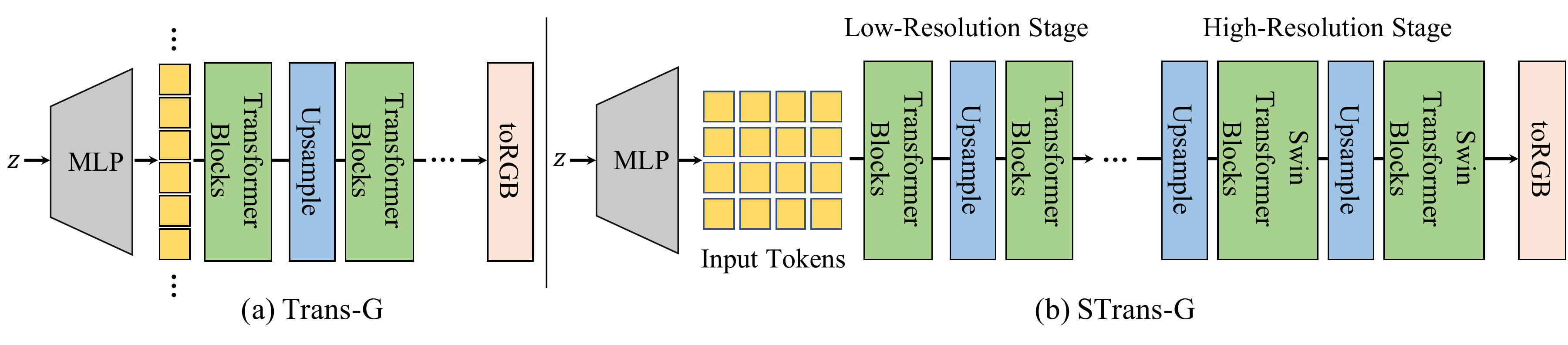}
    \end{center}
    \vspace{-24pt}
    \caption{(a) illustrates the overall structure of the baseline generator, Trans-G, consisting of standard Transformer blocks. (b) An overview of STrans-G that adopts localized attention modules in high-resolution stages.}
    \label{fig:g-arch}
    \vspace{-15pt}
\end{figure*}

\vspace{-3pt}
\section{Methodology}
\label{sec:method}
\vspace{-3pt}

While Transformer has shown impressive performance in representation learning~\cite{vaswani2017attention,xie2021segformer, carion2020end}, early attempts to adopt it for generative modeling mostly lead to dissatisfying results.
In the following, we investigate why the existing models fail to generate high-quality images, and further explore useful practices for designing competitive Transformer-based GANs under both conditional and unconditional settings.

\subsection{Transformer-based Generator}
    

\noindent
\textbf{Trans-G Baseline.}
To design the generator, we start from a straightforward baseline structure, Trans-G, which is composed of standard vision Transformer blocks, as shown in Fig.~\ref{fig:g-arch}(a).
%
Unlike traditional CNN-based models, Trans-G reshapes the spatial feature map $(H_0, W_0, C)$ into a sequence of input tokens $(H_0 \times W_0, C)$.
By stacking upsampling operation and Transformer blocks, Trans-G gradually increases the resolution of the feature map until it meets the target scale $H\times W$.
In such a design, the global attention module in Transformer blocks enhances the modeling ability of long-range dependencies, and several recent studies based on it~\cite{lee2021vitgan,park2021styleformer} present promising results in low-resolution synthesis.

\begin{figure*}[t]
    \begin{center}
    \includegraphics[width=\linewidth]{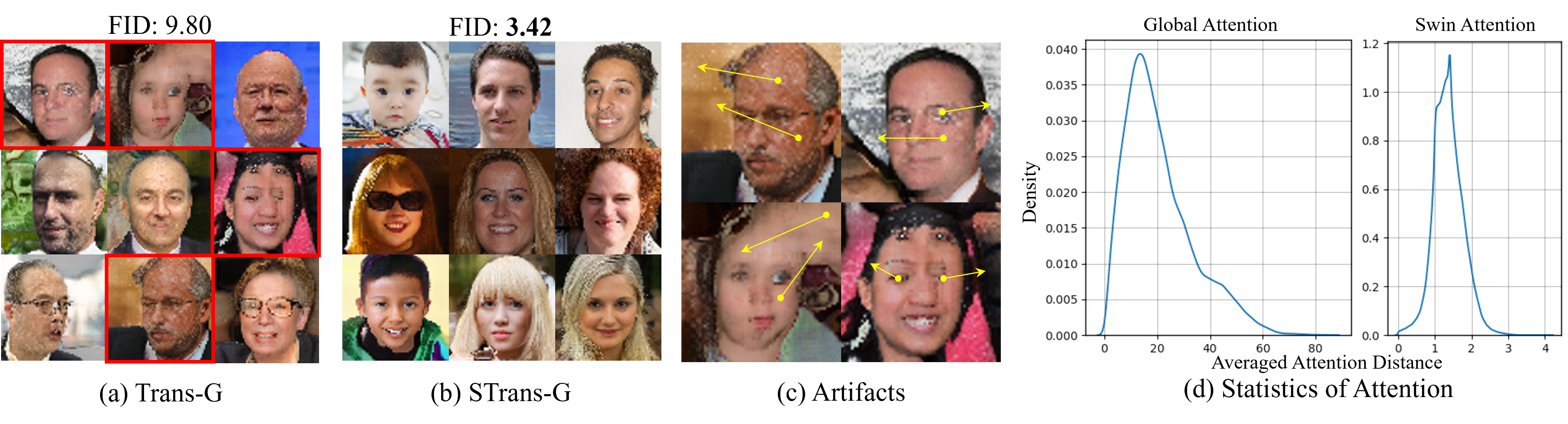}
    \end{center}
    \vspace{-24pt}
    \caption{(a, b) show FFHQ $64\times 64$ samples from Trans-G and STrans-G, respectively. (c) We analyze the failure cases from Trans-G. Given a query pixel at the starting point, the end point of the yellow arrow in (c) represents the pixel with the highest attention score in the last attention layer. (d) We plot the distribution of the averaged attention distance, indicating the spatial distance from the query token position to the location that modules attend to. The statistics are calculated in the last stage of the generator.}
    \label{fig:locality}
    \vspace{-19pt}
\end{figure*}

Samples generated by Trans-G, however, often contain severe artifacts and unrealistic details, leading to poor visual quality (Fig.~\ref{fig:locality}(a)). In particular, we train Trans-G with a strong CNN-based discriminator adopted in StyleGAN2-ADA~\cite{karras2020analyzing,karras2020training} on $64\times 64$ FFHQ\footnote{The $64^2$ FFHQ dataset is downsampled from the original $1024^2$ scale with the bilinear interpolation.} dataset~\cite{karras2019style}.
As shown in Fig.~\ref{fig:locality}(a), while Trans-G can synthesize reasonable global structures in almost all the samples, the generator fails to synthesize realistic high-frequency textures in local regions causing unpleasant artifacts, \eg, the white points in the facial areas.

Given the presence of such high-frequency artifacts, we believe the root cause is the attention modules in the high-resolution stages. 
%
Consequently, we first analyze the last high-resolution attention map closest to the final image feature.
The yellow arrows in Fig.~\ref{fig:locality}(c) point to the pixel with the highest attention score \wrt the query pixel, \ie, the starting point of the arrow.  
As can be observed, the self-attention module in Trans-G aggregates features from an unrelated distant region, \eg, the query pixel in the facial region attends to a background pixel. 
Such a phenomenon suggests the possible pitfall of introducing global attention in the high-resolution feature stage.

We further study the behavior of the self-attention module by computing the average distance between the query token position and the locations it attends to. 
Figure \ref{fig:locality}(d) shows the statistics collected by sampling 10k images.
%
%
%
Without any strict constraints, the global attention module always attends to pixels with a medium distance (10-20 pixels) that cannot offer direct guidance in synthesizing local textures.
Importantly, a significant amount of samples in Trans-G even attend to unreasonably distant pixels ($>$32 pixels) in the highest scale, which can easily result in unpleasant noisy artifacts as shown in Fig.~\ref{fig:locality}(c).
%

    

\noindent
\textbf{STrans-G.}
The analysis above suggests that those Transformer blocks in high-resolution stages should follow the locality nature of image data rather than perform feature reassembly in an unconstrained manner.
The finding motivates us to explore the effect of various local attention mechanisms~\cite{liu2021swin, zhang2021aggregating,jiang2021transgan} in generating realistic high-resolution images.

After trying different local attention modules~\cite{jiang2021transgan,zhao2021improved} extensively, we select the Swin architecture~\cite{liu2021swin} as the building block in high-resolution stages starting from $16\times 16$ pixels (Fig.~\ref{fig:g-arch}(b)).
In the first several stages, since feature maps have a small spatial dimension, we directly apply the standard Transformer block~\cite{dosovitskiy2020image} to capture the global structure.
Swin not only promotes locality with its non-overlapped local windows, but it also helps attention exchange with shifted window attention. In addition, it does not introduce more parameters, and it keeps a linear computational complexity ($\mathcal{O}(HW)$) in various input resolutions. All such designs make the Swin layer an attractive option for a Transformer-based GAN.

After switching to the Swin architecture with a window size of $M = 4$, we observe a significant improvement in both qualitative and quantitative results, as shown in Fig.~\ref{fig:locality}(b).
The unpleasant artifacts disappear and the FID is improved by 65\% over Trans-G.
Figure~\ref{fig:locality}(d) presents the distribution of the attention distance in Swin attention modules from the last stage of the generator.
%
Interestingly, the Swin architecture tends to focus on nearby regions within only a vicinity of two pixels.
%

\noindent \textbf{Relationship with Concurrent Work.}
Several concurrent studies~\cite{zhao2021improved, jiang2021transgan} also introduce local attention mechanisms into GANs.
These prior studies propose local attention mostly from a computational perspective, \ie, to avoid the quadratic complexity of global attention modules. This goal differs significantly from our motivation of reducing artifacts and improving image quality. 
Specifically, TransGAN~\cite{jiang2021transgan} propose the grid self-attention to aggregate features in a $16\times 16$ window. 
However, the grid attention lacks connections between local regions. Information from different local regions thus cannot be effectively communicated, leading to unpleasant results.
While HiT~\cite{zhao2021improved} remedies this issue with dilated attention, it brings $\mathcal{O}((HW)^{1.5})$ computational complexity that is unaffordable for higher resolutions.
Thus, HiT can only apply Transformer blocks in low-resolution stages, while the synthesis in high-resolution stages cannot benefit from the self-attention mechanism.
In contrast, STrans-G uses the shift-window scheme in Swin blocks, allowing effective information propagation between neighboring regions while restricting the computations in local windows.
%
In Sec.~\ref{sec:ablation}, we provide a comprehensive comparison among different local attention mechanisms and a systematic analysis of their effects on image generation.
Note that we do not claim contributions on the Swin architecture. Instead, this study focuses on investigating the benefits of adopting local attention in Transformer-based GANs and the underlying reasons.

\begin{figure*}[t]
    \begin{center}
    \includegraphics[width=0.95\linewidth]{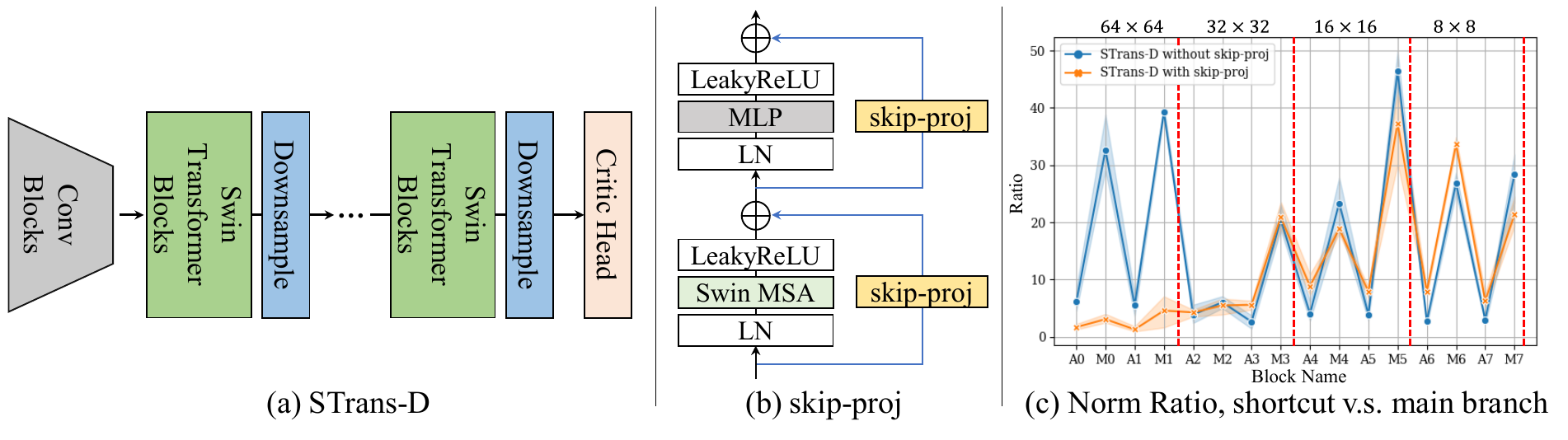}
    \end{center}
    \vspace{-25pt}
    \caption{(a) The overall architecture of STrans-D. In (b), we illustrate the detailed architecture of the Swin Transformer block with \bfit{skip-proj} in STrans-D. (c) presents the norm ratio between the shortcut (blue path in (b)) and the main branch that contains self-attention and MLP blocks. `Ai' and `Mj' denotes the i-th attention block and the j-th MLP block, respectively. 
    The red vertical line in (c) represents the position of a downsampling operation. At the top of (c), we offer the resolution of the features in the current stage. We train each configuration with three different random seeds. The error bar on each curve represents the variance across different runs. }
    \label{fig:strans-d}
    \vspace{-16pt}
\end{figure*}

\subsection{Transformer-based Discriminator}
\label{sec:strans-d}

%
Similar to STrans-G, we also use Swin layers for the discriminator, termed as STrans-D. 
Nevertheless, directly applying Swin in the discriminator leads to unstable adversarial training in our experiments, which often results in degenerated image qualities.
Next, we summarize our experience to improve the baseline model of STrans-D and present a novel skip-projection scheme as a remedy.

\noindent \textbf{Empirical Strategies for STrans-D.}
First, instead of starting with a patch embedding module as in most vision Transformers~\cite{dosovitskiy2020image}, we adopt a lightweight convolutional block to downsample the original input by $4\times$ and project the image tensor to an arbitrary dimension.
Compared to patch embedding, the convolutional token extractor adopts overlapped patches, keeping more detailed information.
Second, we adopt the equalized learning rate~\cite{karras2017progressive} in all attention modules and MLPs.
This is motivated by the slow and unsatisfactory convergence of Transformer blocks in the discriminator when they use a small learning rate to stabilize its training.
We resolve this issue by setting a large learning rate for the whole discriminator and introducing a special scaler to multiply the learnable parameters from Transformer blocks at runtime.
Additionally, we replace GeLU~\cite{hendrycks2016gaussian} with LeakyReLU~\cite{maas2013rectifier} and add a nonlinear activation function at the end of the attention and MLP blocks.
We note that while many empirical strategies already exist in separate works~\cite{lee2021vitgan,karras2017progressive}, it requires substantial efforts to identify the most effective components from the literature and further combine them to build a strong baseline that will well benefit the field.
%
More details about these improvements can be found in the supplementary material.



\noindent
\textbf{Skip-Projection.}
Despite the empirical improvements above, there is still a significant performance gap between the Transformer-based STrans-D and the discriminator of StyleGAN2.
To solve this problem, we perform a careful analysis of STrans-D, and find it is the residual connections in Transformer blocks that lead to an undesirable information flow, making the feedback from the discriminator degrades.
In particular, we plot the norm ratio~\cite{raghu2021vision} of the residual connections, which is defined as $||x|| / ||f(x)||$, where $x$ denotes the features from the shortcut and $f(x)$ represents the transformation of $x$ from the main branch.
The norm ratio is collected from the checkpoint with the best FID and thus faithfully reflects the information flow in the network.

As shown in Fig.~\ref{fig:strans-d}(c), STrans-D (blue curve) has an extremely high ratio in the $64\times 64$ stage, indicating that the information almost flows through the shortcuts instead of the sub-layers containing the self-attention and MLP blocks.
It is from $32\times 32$ stage that STrans-D starts processing features with the main branch.
%
%
This deviates from our expectation that a discriminator should have paid more attention to the high-resolution features containing plentiful details, so that its feedback can better guide high-quality synthesis.

To avoid such an unhealthy behavior, we apply a skip-projection layer, termed as \bfit{skip-proj}, which performs a linear projection in the residual connection (Fig.~\ref{fig:strans-d}(b)). 
The linear projection layer adaptively adjusts the scale of features in residual connections, allowing the discriminator to experience a more reasonable information flow.
The comparison in Fig.~\ref{fig:strans-d}(c) shows that skip-proj prevents the escalation of the norm ratios in early stages.
Benefiting from the high-resolution features, STrans-D with skip-proj effectively reduces the gap against the StyleGAN2 discriminator. 
It is noteworthy that even with the proposed skip-proj, the discriminators still learn to rely on the shortcut branch instead of the main branch in low-resolution stages like $16\times 16$.
This phenomenon suggests that the low-resolution features contain less informative content that can benefit the discrimination task in high-quality synthesis.
In Sec.~\ref{sec:exp-strans-d}, we show how STrans-D evolves step by step and verify the effectiveness of skip-proj in speeding up convergence and improving final performance. 

\begin{figure*}[t]
    \begin{center}
    \includegraphics[width=0.9\linewidth]{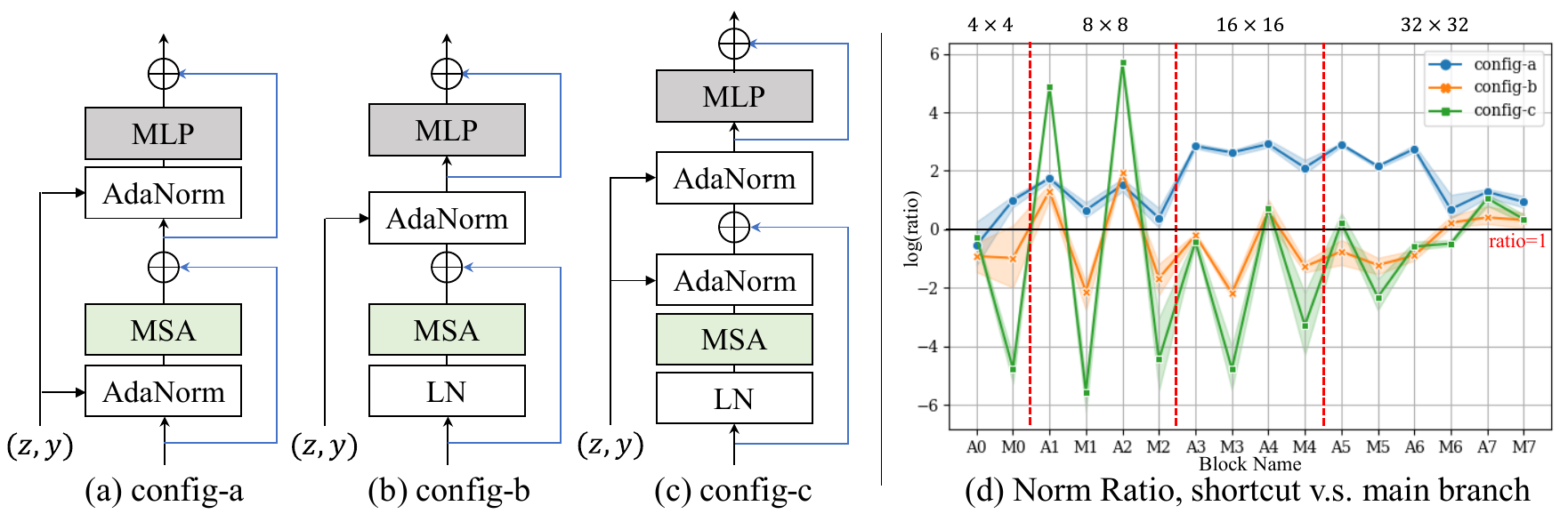}
    \end{center}
    \vspace{-24pt}
    \caption{(a)-(c) show three alternative designs of adopting AdaNorm in a Transformer block. (d) presents the norm ratio between the shortcut (blue path in (a)-(c)) and the main branch. `Ai' and `Mj' denote the i-th attention block and the j-th MLP block, respectively.
    The red vertical line in (d) indicates the position of an upsampling operation. At the top of (d), we offer the scale of the features in the current stage. We train each configuration with three random seeds. The error bar on each curve represents the variance across different runs.}
    \label{fig:c-strans-g}
    \vspace{-18pt}
\end{figure*}

\vspace{-1pt}
\subsection{Conditional Image Generation}
\vspace{-1pt}

We also study Transformer-based GAN for conditional image generation.
%
%
Related to our previous discussions on skip-proj, we find that the direct shortcut in Transformer blocks also has a profound influence on the design of conditional generators. 

In convolutional GANs, the category condition is typically injected by AdaNorm on the main branch~\cite{dumoulin2016learned,huang2017arbitrary,park2019SPADE}:
{
\begin{equation}
    \label{eq:adanorm}
    \mathrm{AdaNorm}(x, y, z) = \gamma (y, z) \cdot \mathrm{Norm}(x) + \beta (y, z),
\end{equation}
}
\hspace{-2.4pt}where $x$, $y$ and $z$ are input features, category conditions, and noise vectors respectively.
The $\mathrm{Norm}(\cdot)$ can be realized with different normalization layers, \eg, Layer normalization~\cite{ba2016layer}, Batch Normalization~\cite{ioffe2015batch}, and Instance Normalization~\cite{ulyanov2016instance} (correspondingly, the AdaNorm is called as AdaLN, AdaBN, and AdaIN).
%
%
$\gamma(\cdot)$ and $\beta(\cdot)$ modulate the normalized feature $\mathrm{Norm}(x)$ according to the conditional label $y$ and noise vector $z$.
%


Our attempt of adopting AdaLN directly in Transformer blocks (config-a in Fig.~\ref{fig:c-strans-g}(a)) fails. In particular, we find the FID stops decreasing in the early stage.
To trace the source of failure, we plot the norm ratio of this baseline configuration. 
As shown in Fig.~\ref{fig:c-strans-g}(d), there exist several blocks with a large norm ratio, indicating that some AdaNorm layers in the main branch contribute little to the intermediate features, leading to a loss of conditional information.
%
%
To guarantee the injection of conditional information, a simple solution is to apply the AdaNorm in the trunk, as shown in config-b of Fig.~\ref{fig:c-strans-g}(b).
In this way, the features from both the shortcut and the MLP branch are ensured to contain class information. 
Interestingly, config-b brings the norm ratio below the baseline config-a.

Config-b, however, is not a stable solution as it causes undesirable mode collapse in several training cases.
Since mode collapse reflects a failure in modeling global structures, we focus on the early attention blocks which compute global attention.
As shown in Fig.~\ref{fig:c-strans-g}(c), after adopting another AdaNorm at the end of the attention block (config-c), we find mode collapse disappears in the early stage.
As shown in Fig.~\ref{fig:c-strans-g}(d), config-c leads to an extremely high ratio in the second and third attention blocks, hence avoiding collapsed global structures from the main branch.
We hypothesize this is mainly due to $\gamma(\cdot)$ and $\beta(\cdot)$ in Eq.~(\ref{eq:adanorm}), which can serve as a gating function to mask undesirable signals.
Meanwhile, the lower norm ratio in config-c (green curve in Fig.~\ref{fig:c-strans-g}(d)) suggests that the model learns to depend more on the MLP blocks to mitigate the impacts of bypassing the global attention modules.

In summary, Fig.~\ref{fig:c-strans-g}(c) depicts the structure that we found useful for adopting AdaNorm in Transformer blocks. We call it as \bfit{AdaNorm-T}, and correspondingly AdaIN-T/AdaBN-T if IN/BN is used. Note that the AdaNorm after the attention modules is only adopted in the global attention blocks, as we find applying it in Swin blocks cannot further improve the performance. 
The effectiveness of AdaNorm-T is verified on the CIFAR10 and \textsc{ImageNet} datasets in Sec.~\ref{sec:exp-strans-g}.
StyleFormer~\cite{park2021styleformer} also discusses the role of layer normalization. However, their modification of adding another normalization layer aims to stabilize the attention module instead of injecting conditional priors.
Besides, the layer-by-layer fluctuation in Fig.~\ref{fig:strans-d}(c) and Fig.~\ref{fig:c-strans-g}(d) is also observed in popular vision Transformers~\cite{raghu2021vision}, which may inspire the future architectural design.
%

%

\vspace{-3pt}
\section{Experiments}
\label{sec:exp}
\vspace{-3pt}

\noindent
\textbf{Implementation Details.}
To reduce computational costs, we set the channel expansion ratio to 2 for the MLP modules in Transformer blocks.
The input token dimension is 512. We adopt four attention heads by default.
We select Adam~\cite{kingma2014adam} optimizer with $\beta_1=0, \beta_2=0.99$ to train our models.
STrans-G and STrans-D are optimized with a learning rate of 0.0001 and 0.002, respectively. 
All of our models are trained on 8 Tesla V100 GPUs in PyTorch~\cite{NEURIPS2019_9015}.
We evaluate our models with FID and Inception Score (IS)~\cite{salimans2016nips}, following the implementation in StyleGAN2-ADA~\cite{karras2020training}.
More details about models and training pipelines can be found in the following sections and supplementary material.


\vspace{-1pt}
\subsection{STrans-G}
\label{sec:exp-strans-g}
\vspace{-1pt}

\begin{table*}[t]
    \caption{(a, b) present the comparison of FID with popular unconditional methods\protect\footnotemark. In conditional generation, (c, d) show the comparison with CNN-based conditional models. STrans-G in FFHQ $256^2$ only contains \textbf{20M} parameters, while StyleGAN2 and HiT have 30M and 46M parameters in the generator. $\downarrow$ indicates lower the better, and $\uparrow$ indicates higher the better.}
    \vspace{-5pt}
    \label{tab:main-res}
    \vspace{-5pt}
    \begin{subtable}[h]{0.22\textwidth}
        \centering
        \caption{\textsc{CelebA} $64^2$}
        \vspace{-4pt}
        \begin{adjustbox}{width=\textwidth,center}
        \begin{tabular}{|c|c|}
            \hline
          \textbf{Method}  & FID $\downarrow$ \\ \hline \hline
           HDCGAN~\cite{curto2017high} & 8.44 \\
           COCOGAN~\cite{lin2019coco} & 4.00 \\
           NCP-VAE~\cite{aneja2020ncp} & 5.25 \\ 
           StyleGAN2-ADA~\cite{karras2020training} & 3.16 \\
           UDM~\cite{kim2021score} & 2.78 \\
           TransGAN~\cite{jiang2021transgan} & 5.01 \\ \hline
            STrans-G & \textbf{2.03} \\ \hline
           \end{tabular}
        \end{adjustbox}
       \label{tab:celeba}
    \end{subtable}
    \hfill
    \begin{subtable}[h]{0.22\textwidth}
        \centering
        \caption{FFHQ $256^2$}
        \vspace{-4pt}
        \begin{adjustbox}{width=\textwidth,center}
        \begin{tabular}{|c|c|}
            \hline
          \textbf{Method}  & FID $\downarrow$ \\ \hline \hline
           U-NetGAN~\cite{schonfeld2020u} & 7.63 \\
           VQGAN~\cite{esser2021taming} & 11.40 \\
            INR-GAN~\cite{skorokhodov2021adversarial} & 9.57 \\
            CIPS~\cite{anokhin2021image} & 4.38 \\
           StyleGAN2-ADA~\cite{karras2020training} & 3.62 \\
           HiT~\cite{zhao2021improved} & \textbf{2.95} \\ \hline
            STrans-G & 4.84 \\ \hline
           \end{tabular}
        \end{adjustbox}
       \label{tab:ffhq}
    \end{subtable}
    \hfill
    \begin{subtable}[h]{0.28\textwidth}
        \centering
        \caption{CIFAR10 $32^2$}
        \vspace{-4pt}
        \begin{adjustbox}{width=\textwidth,center}
        \begin{tabular}{|c|c|c|}
            \hline
          \textbf{Method}  & FID $\downarrow$ & IS $\uparrow$ \\ \hline \hline
           ProjGAN~\cite{miyato2018cgans} & 17.50 & 8.62 \\
           BigGAN~\cite{brock2018large} & 14.73 & 9.22 \\
           MultiHinge~\cite{kavalerov2019cgans} & 6.40 & 9.68 \\
           FQGAN~\cite{zhao2020feature} & 5.59 & 8.48 \\
           Mix-MHingeGAN~\cite{tang2020lessons} & 3.60 & 10.21 \\
           StyleGAN2-ADA~\cite{karras2020training}  & \textbf{2.42} & 10.14 \\ \hline
            STrans-G & 2.77 & \textbf{11.62} \\ \hline
           \end{tabular}
        \end{adjustbox}
       \label{tab:cifar10}
    \end{subtable}
    \hfill
    \begin{subtable}[h]{0.239\textwidth}
        \centering
        \caption{\textsc{ImageNet} $128^2$}
        \vspace{-4pt}
        \begin{adjustbox}{width=\textwidth,center}
            \begin{tabular}{|c|c|c|}
                \hline
              \textbf{Method}  & FID $\downarrow$ & IS $\uparrow$ \\ \hline \hline
               ProjGAN~\cite{miyato2018cgans} & 27.62 & 36.80\\
               SAGAN~\cite{zhang2019self} & 18.65 & 52.52 \\
               YLG~\cite{daras2020your} & 15.94 & 57.22 \\
               ContraGAN~\cite{kang2020contragan} & 19.69 & 31.10 \\
               FQ-GAN~\cite{zhao2020feature} & 13.77 & 54.36 \\
               BigGAN~\cite{brock2018large} & \textbf{8.70} & \textbf{98.80} \\ \hline
                STrans-G  & \imgnetfid & \imgnetis \\ \hline
               \end{tabular}
        \end{adjustbox}
       \label{tab:imgnet}
    \end{subtable}
    \vspace{-6pt}
\end{table*}

\noindent
\textbf{Setup.}
To evaluate the effectiveness of STrans-G, we first adopt strong CNN-based discriminators to pair with STrans-G in both unconditional and conditional settings.
For experiments in FFHQ $256^2$~\cite{karras2019style} and CIFAR10~\cite{krizhevsky2009learning}, we use the discriminator proposed in StyleGAN2-ADA~\cite{karras2020training}, and follow its training configuration.
Note that the path regularization loss is removed during optimizing STrans-G to reduce the training costs.
For \textsc{CelebA}~\cite{liu2015faceattributes}, we adopt the same training settings with FFHQ $256^2$.
In the experiments for \textsc{ImageNet}~\cite{krizhevsky2012imagenet}, as we find the StyleGAN2 discriminator always leads to an early stopping in training with such large-scale data, we switch to the discriminator proposed by \cite{brock2018large} and follow its setting with a batch size of 2048.
%
%
To efficiently inject class information, we adopt AdaIN-T in CIFAR10 and AdaBN-T in \textsc{ImageNet}.

\begin{figure*}[t]
    \begin{center}
    \includegraphics[width=0.99\linewidth]{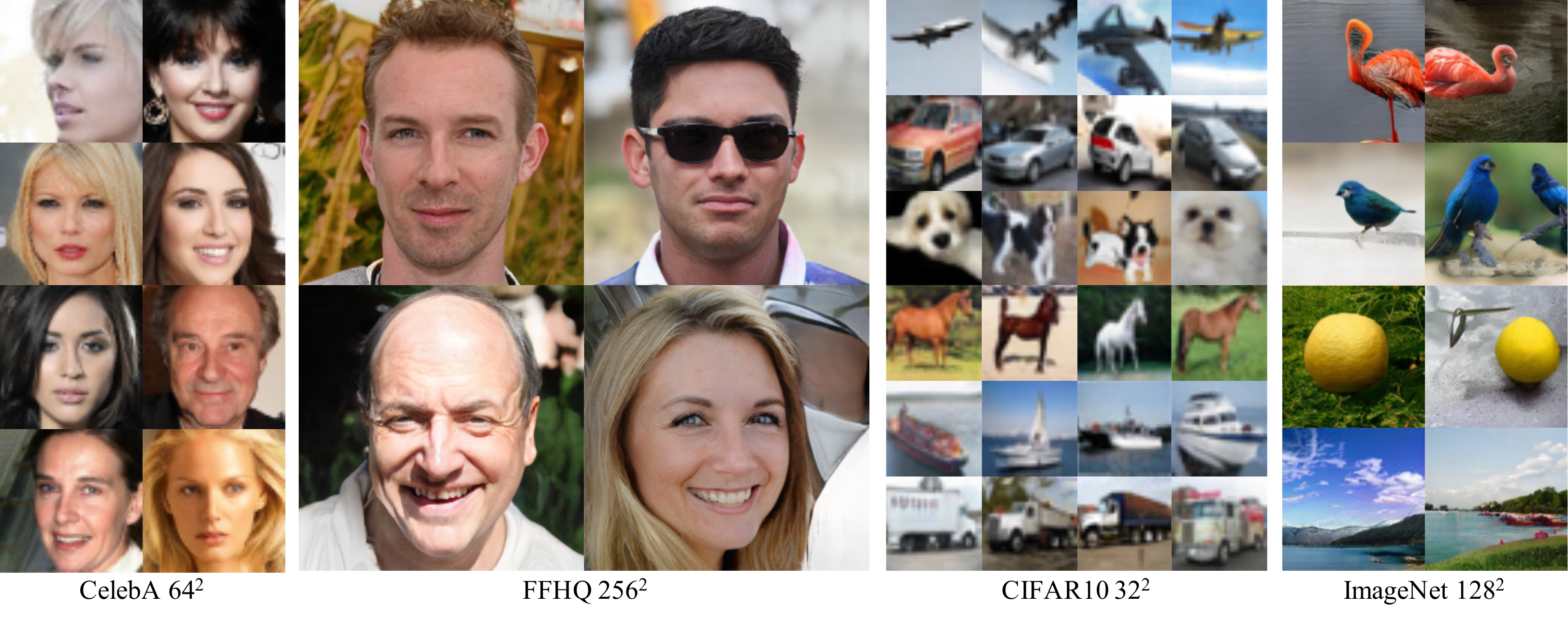}
    \end{center}
    \vspace{-24pt}
    \caption{Exemplar images generated from STrans-G in unconditional and conditional settings.}
    \vspace{-18pt}
    \label{fig:samples}
\end{figure*}

\noindent
\textbf{Results.}
In Tab.~\ref{tab:main-res}, we compare STrans-G with state-of-the-art CNN-based and Transformer-based models.
In unconditional generation, STrans-G significantly outperforms all previous methods in \textsc{CelebA} $64^2$.
It also achieves competitive performance in the high-resolution setting of FFHQ $256^2$.
It is noteworthy that the gap between HiT~\cite{zhao2021improved} and STrans-G is mainly caused by the model size and large training batches.
In fact, STrans-G only contains \textbf{20M} parameters, while StyleGAN2 and HiT have 30M and 46M parameters in the generator, respectively.  
In addition, HiT applies a large batch size of 256 using TPUs, while STrans-G's batch size is 64 by default.

For conditional image generation, with the proposed AdaIN-T layer, STrans-G improves the SOTA Inception Score (IS)~\cite{salimans2016nips} from 10.14 to 11.62 on CIFAR10, as shown in Tab.~\ref{tab:main-res}(c). 
Since CIFAR10 is widely adopted as a limited data benchmark~\cite{karras2020training}, this result also suggests the robustness of STrans-G in modeling real distributions with limited data. 
In the \textsc{ImageNet} evaluation, STrans-G achieves superior performance over current popular methods except for the extensively tuned BigGAN~\cite{brock2018large}. 
The gap between STrans-G and the SOTA BigGAN model suggents room for improvement in Transformer-based GAN.
Nonetheless, for the first time, this study shows the potential of Transformers in the conditional generation.
%
%
Figure \ref{fig:samples} presents unconditional and conditional samples generated from STrans-G.
The visual quality of these generated samples suggests the great potential of using pure Transformer blocks in GANs.
Additional qualitative results, including interpolation in the latent space and discussion abount runtime performance, can be found in the supplementary material.

\subsection{STrans-D}
\label{sec:exp-strans-d}

\newcommand{\disctab}{
    \begin{tabular}{|l|c|c|}
        \hline
        \textbf{Configuration}  & FFHQ & \textsc{CelebA} \\ \hline \hline
        a. Swin Baseline &20.19 &14.63\\
        b. + Conv Extractor   &14.10&11.87\\
        c. + Equalized LR &8.95&6.13\\
        d. + LeakyReLU      &8.63 &4.44\\
        e. + skip-proj    &5.98&3.41\\ \hline \hline
        f. StyleGAN2-D &4.84&2.03\\ \hline
    \end{tabular}
}

\begin{figure*}[t]
    \footnotesize%
    \newcommand{\h}{0.4\linewidth}%
    \newcommand{\hh}{0.295\linewidth}%
    \newcommand{\hhh}{0.001\linewidth}%
    \newcommand{\vvnew}{30mm}%
    \newcommand{\vvv}{1.}%
    \subtabletopmod{\h}{\vvnew}{\vvv}{\disctab}{-2pt}\hspace{-2pt}%
    \parbox[b]{\hh}{
    \includegraphics[width=\linewidth]{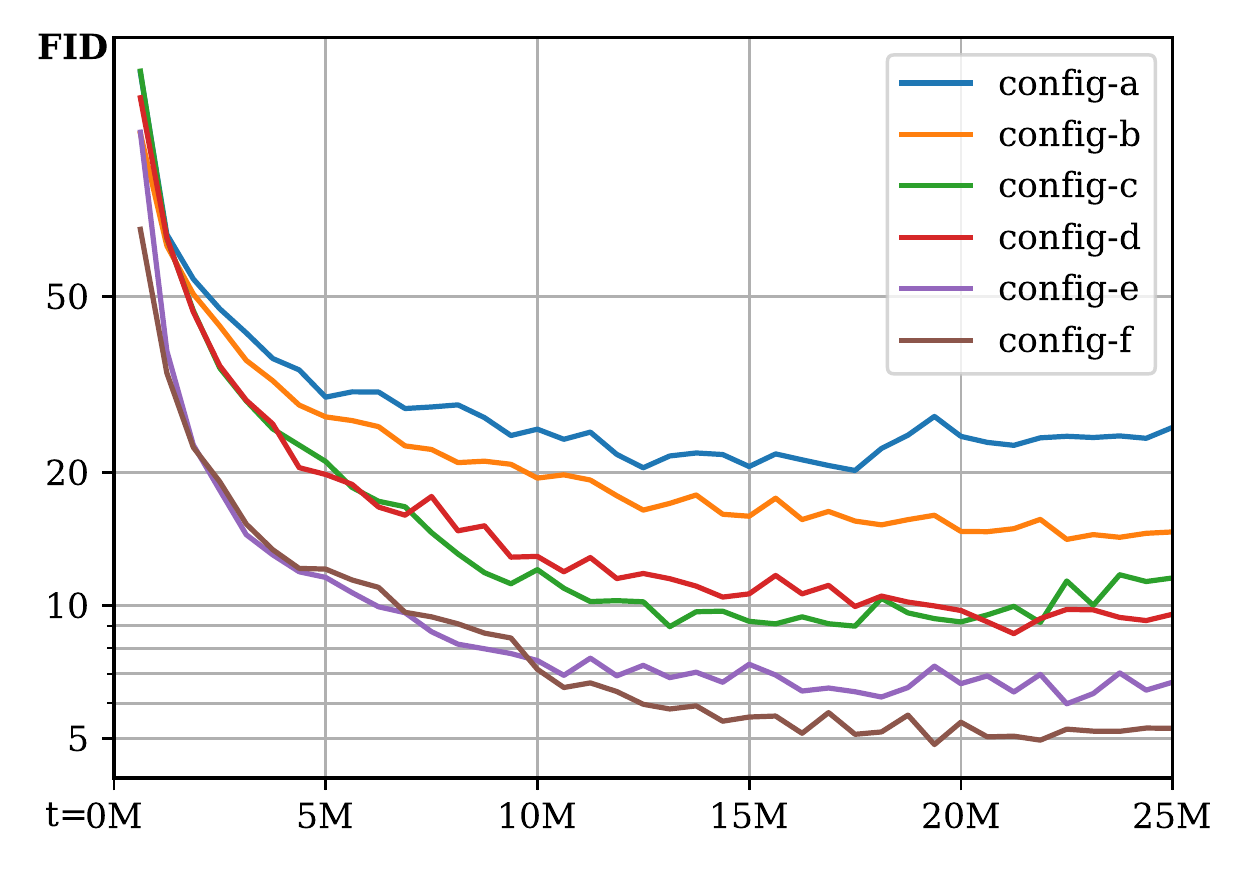}}%
    \hspace{-2pt}%
    \parbox[b]{0.295\linewidth}{
    \includegraphics[width=\linewidth]{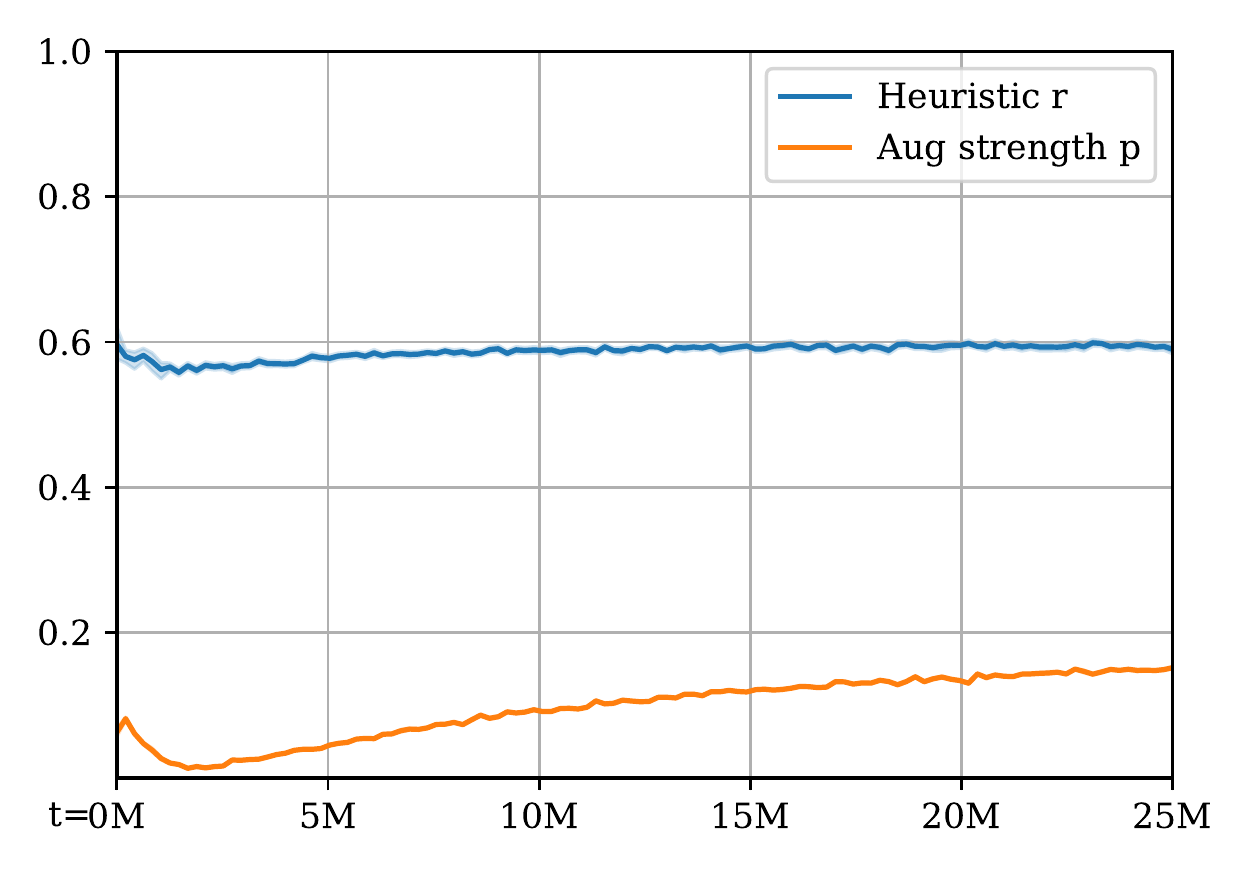}} \\ \vspace{-15pt} \\
    \makebox[\h][c]{(a) FID}\hspace{-2pt}%
    \makebox[0.3\linewidth][c]{(b) Training in FFHQ}\hspace{-5pt}%
    \makebox[0.3\linewidth][c]{(c) Evolution of $r$ and $p$ in config-e}%
    \vspace{-8pt}
    \caption{
    (a) The FID for various discriminator configurations described in Sec.~\ref{sec:strans-d}.
    (b) We plot the evaluation FID during the training stage for each configuration in (a).
    (c) In FFHQ $256^2$, we present the evolution of two indicators proposed in ADA~\cite{karras2020training}, \ie, overfitting heuristic $r$, and augmentation strength $p$.
    }
    \vspace{-18pt}
    \label{fig:disc-exp}
\end{figure*}

\footnotetext{Since ViTGAN~\cite{lee2021vitgan} and StyleFormer~\cite{park2021styleformer} use the aligned version of \textsc{CelebA}, we cannot directly compare with them.}

\noindent
\textbf{Setup.}
Here, we study the effectiveness of STrans-D by pairing it with the unconditional STrans-G.  
The conditional discriminator can be easily implemented by adding conditional projection~\cite{miyato2018cgans} in the critic head (Fig.~\ref{fig:strans-d}(a)), which is not relevant to the Transformer backbone.
Thus, in this work, we only conduct experiments in the unconditional generation. The training schedule is kept the same as the StyleGAN2 discriminator (Sec.~\ref{sec:exp-strans-g}) to guarantee a fair comparison.
More implementation details about STrans-D can be found in supplementary material.

\noindent
\textbf{Results.}
Fig.~\ref{fig:disc-exp}(a) shows how we improve the initial Swin-based discriminator baseline step by step.
Besides, Fig.~\ref{fig:disc-exp}(b) presents the evaluation FID over the training process, so that we can conveniently compare the convergence speed.

As the original patch embedding module in vision Transformer takes non-overlapped patches as input, it is not efficient in extracting input tokens.
Switching to a convolutional feature extractor can bring a clear improvement of FID in Fig.~\ref{fig:disc-exp}(a).
We then adopt the equalized learning rate to adjust the learning rate for the Transformer blocks at runtime.
Thanks to this scheme, we can train the discriminator with a large learning rate, speeding up the convergence and decreasing the FID.
In addition, adopting LeakyReLU also slightly improves the performance.
%
%
Finally, the proposed skip-proj significantly reduces this gap (variant-e in Fig.~\ref{fig:disc-exp}(a)) and accelerates the convergence in the early stage (Fig.~\ref{fig:disc-exp}(b)).

We highlight that STrans-D is not prone to the overfitting problem.
As introduced in StyleGAN2-ADA~\cite{karras2020training}, the overfitting of a discriminator can be characterized by two indicators: the overfitting heuristic $r$ and augmentation strength $p$. 
If a discriminator meets the overfitting problem, $p$ should increase quickly in training and ends up with a value larger than 0.2, while $r$ should reach a value close to 1.
We plot the evolution of $r$ and $p$ over the course of training with variant-e in Figure \ref{fig:disc-exp}(c).
The heuristic $r$ is always below 0.6 and the augmentation strength $p$ ends up with a small probability, suggesting that STrans-D has not met the overfitting problem. We further discuss the effects of differentiable data augmentation in supplementary material.

\subsection{Further Analysis}
\label{sec:ablation}

\newcommand{\attentable}{\begin{tabular}{|l|c|c|cc|}
    \hline
    \multirow{2}{*}{\textbf{Attention}} & \textsc{CelebA} & FFHQ & \multicolumn{2}{c|}{CIFAR10} \\
                               & FID$\downarrow$    & FID$\downarrow$  & FID$\downarrow$                    & IS$\uparrow$  \\ \hline \hline
    Trans                      &  3.77  &   --   & \multicolumn{1}{c|}{4.01}  &  9.25   \\
    +Rel. Pos.                 &  3.70  &   --   & \multicolumn{1}{c|}{4.25}  &  9.11   \\ \hline\hline
    Grid                       &   3.63     &    6.52  & \multicolumn{1}{c|}{3.82}  &  9.31   \\
    MultiAxis                  &    3.60    &  5.63    & \multicolumn{1}{c|}{3.22}  & 10.12  \\
    Swin                       &  \textbf{2.03}  & \textbf{4.84} & \multicolumn{1}{c|}{\textbf{2.77}}  & \textbf{11.62}    \\ \hline
    \end{tabular} 
}

\newcommand{\adanormtable}{\begin{tabular}{|c|c|c|c|}
    \hline
    \textbf{Architecture}              & Norm & FID$\downarrow$ & IS$\uparrow$ \\ \hline \hline
    \multirow{2}{*}{config-a} & LN   &  4.94  & 9.53   \\
                              & IN   &  5.78 &  8.13  \\ \cline{1-2}
    \multirow{2}{*}{config-b} & LN   &   13.30  &  5.27  \\
                              & IN   &  10.14   &   5.48 \\ \cline{1-2}
    \multirow{2}{*}{config-c} & LN   & 3.99    & 9.72   \\
                              & IN   &2.77&11.62\\ \hline
    \end{tabular}
}

\newcommand{\adatable}{\begin{tabular}{|l|c|cc|}
    \hline
    \multirow{2}{*}{\textbf{Dataset}} & \multirow{2}{*}{Disc} & \multicolumn{2}{c|}{ADA} \\
                             &                       & w/o         & w/         \\ \hline \hline
                             \multirow{2}{*}{\textsc{CelebA}}                & CNN             &  2.41     &     2.03     \\
                             & Trans             & 3.32&3.16\\ \hline
    \multirow{2}{*}{FFHQ}    & CNN             &    5.24    &  4.84      \\
                             & Trans             &   6.22   & 5.98  \\ \hline
    CIFAR10                  & CNN             &     4.07       &  2.77           \\ \hline
    \end{tabular}   
}

\begin{table}[t]
    \centering
    \caption{
    We study the effects of adopting different attention mechanisms in STtrans-G. `Trans' indicates the standard self-attention module~\cite{dosovitskiy2020image}, while `Grid'~\cite{jiang2021transgan} and `MultiAxis'~\cite{zhao2021improved} are recent studies applying the localized idea in attention modules. For a fair comparison, we also add relative positional encoding~\cite{liu2021swin} for the standard Transformer blocks, denoted by `+ Rel. Pos.'.}
    \vspace{-7pt}
    \label{tab:attntype}
    \begin{adjustbox}{width=0.73\linewidth,center}
    \begin{tabular}{|l|c|c|cc|}
        \hline
        \multirow{2}{*}{\textbf{Attention}} & \textsc{CelebA} & FFHQ & \multicolumn{2}{c|}{CIFAR10} \\
                                   & FID$\downarrow$    & FID$\downarrow$  & FID$\downarrow$                    & IS$\uparrow$  \\ \hline \hline
        Trans                      &  3.77  &   --   & \multicolumn{1}{c|}{4.01}  &  9.25   \\
        + Rel. Pos.                 &  3.70  &   --   & \multicolumn{1}{c|}{4.25}  &  9.11   \\ \hline\hline
        Grid                       &   3.63     &    6.52  & \multicolumn{1}{c|}{3.82}  &  9.31   \\
        MultiAxis                  &    3.60    &  5.63    & \multicolumn{1}{c|}{3.22}  & 10.12  \\
        Swin                       &  \textbf{2.03}  & \textbf{4.84} & \multicolumn{1}{c|}{\textbf{2.77}}  & \textbf{11.62}    \\ \hline
        \end{tabular} 
    \end{adjustbox}
    \vspace{-15pt}
\end{table}

\noindent
\textbf{Different Local Attention Mechanisms.}
STrans-G is a general framework where we can switch to various attention mechanisms. Here, we study the effects of different possible choices based on the framework of STrans-G.
Table \ref{tab:attntype} presents the influence of different attention mechanisms on the final performance.
With global attention, the standard Transformer block consistently yields inferior image qualities, compared to the others with local attention mechanisms.
Grid attention~\cite{jiang2021transgan} computes self-attention within non-overlapped windows but lacks an efficient way to bridge each local region.
MultiAxis~\cite{zhao2021improved} adopts a sparsed global attention to enhance communication among local patches.
Nevertheless, its design breaks the locality in high-resolution stages while introducing heavy computational costs.
Thus, these models do not give satisfying results as shown in Tab.~\ref{tab:attntype}, and we select the Swin architecture as our default choice of the attention mechanism for high-fidelity image synthesis.

\noindent
\textbf{Ablation Study on AdaNorm-T.}
In Tab.~\ref{tab:adanorm}, we study the effects of different architectural choices and normalization layers in AdaNorm-T.
As shown in Fig.~\ref{fig:c-strans-g}(a), config-a only injects conditional information in residual blocks, which may be ignored by the direct shortcuts.
Thus, it often causes inferior results compared with the config-c in Tab.~\ref{tab:adanorm}.
Wile config-b (Fig.~\ref{fig:c-strans-g}(b)) guarantees the injection of conditional information, it often leads to mode collapse in the early training stages, resulting in extremely high FID.
By adopting an additional AdaNorm at the end of the attention blocks, AdaNorm-T (Fig.~\ref{fig:c-strans-g}(c)) avoids early stopping in training and achieves much better performance (config-c in Tab.~\ref{tab:adanorm}).
Besides, the choice of normalization layers also influences the quality of conditional generation.
We note that the choice of normalization layers highly depends on the dataset. More discussions are provided in the supplementary material.

\begin{table}[t]
    \centering
    \caption{
        We compare different architectural choices and normalization layers in AdaNorm-T. The architectures are depicted in Fig.~\ref{fig:c-strans-g}(a-c).}
    \label{tab:adanorm}
    \vspace{-8pt}
    \begin{adjustbox}{width=0.75\linewidth,center}
        \begin{tabular}{|c|c|c|c|}
            \hline
            \textbf{Architecture}              & Norm & FID$\downarrow$ & IS$\uparrow$ \\ \hline \hline
            \multirow{2}{*}{config-a} & Layer Norm   &  4.94  & 9.53   \\
                                      & Instance Norm   &  5.78 &  8.13  \\ \cline{1-2}
            \multirow{2}{*}{config-b} & Layer Norm   &   13.30  &  5.27  \\
                                      & Instance Norm  &  10.14   &   5.48 \\ \cline{1-2}
            \multirow{2}{*}{config-c} & Layer Norm  & 3.99    & 9.72   \\
                                      & Instance Norm &2.77&11.62\\ \hline
            \end{tabular} 
    \end{adjustbox}
    \vspace{-17pt}
\end{table}


%

%

\section{Conclusion and Limitation}

%
We have conducted a comprehensive empirical study on adopting Transformer in GANs. Through studying various properties of self-attention layers, we observed the crucial role of feature locality in high-quality image synthesis. Swin layer, the popular attention block for classification network, shows promising potential in Transformer-based GAN.
Our analysis also revealed the residual connections, which were assumed to be `safe and useful', could harm the performance of Transformer-based discriminators and conditional generators.
The resulting STransGAN achieves high-quality image generation on a wide variety of benchmarks, which can serve as a simple yet strong baseline for future research.
This work also demystifies the early failures of using Transformers for GANs and paves the promising direction of Transformer-based generative modeling.

Similar to existing works~\cite{park2021styleformer,lee2021vitgan,jiang2021transgan},
one limitation of this study is the missing results in higher resolutions, like $1024\times 1024$, and large models. Since the attention scheme contains computation-unfriendly reshaping or window partition operations, training such models in extremely high resolutions will take unaffordable computational costs. We regard this challenging problem as a future work, where we hope to further reduce the training cost or conduct experiments with sufficient resources.
Another limitation in this study is the gap between STrans-D and the StyelGAN2 discriminator. We have already provided a competitive baseline for future research. Investigating the reason for this gap will help us to understand the difference between CNN-based and Transformer-based models.


{\small
\bibliographystyle{ieee_fullname}
\bibliography{egbib}
}

\clearpage

\appendix

\section{Implementation Details and Discussion}
\label{app:impl}

In this section, we present the implementation details of STransGAN and offer discussions about several details.
All of our models are trained on 8 Tesla V100 GPUs in PyTorch~\cite{NEURIPS2019_9015}.
The models and related codes will be made publicly available.

\setcounter{subsection}{-1}

\subsection{Preliminary for Swin Attention}
Swin Transformer~\cite{liu2021swin} adopts the window partition and only calculates the attention map within each window. Importantly, their design of shifting the window partition between consecutive self-attention layers efficiently bridges the local windows of the preceding layer. In a word, the formulation of this attention mechanism is:
\begin{align}
    \mathbf{h}^l &= \textrm{W-MSA}(\textrm{LN}(\mathbf{h}^{l-1})) + \mathbf{h}^{l-1}, \notag\\
    \label{eq:swin}
    \mathbf{h}^{l+1} &= \textrm{SW-MSA}(\textrm{LN}(\mathbf{h}^{l})) + \mathbf{h}^{l}, 
\end{align}
where $\mathbf{h}^l$ indicates intermediate feature map from the $l$-th layer. $\textrm{W-MSA}$ and $\textrm{SW-MSA}$ denote the window attention and shift-window attention layer, respectively. $\textrm{LN}$ represents the layer normalization layer. Note that we ignore the MLP right after each attention layer in Eq.~\eqref{eq:swin} for the sake of simplicity. Thanks to the design of the shift window, Swin attention layers can model the long-range relationship while reducing the computational complexity to $\mathcal{O}(HW)$.
\subsection{STrans-G}
\label{app:impl-strans-g}

\begin{figure}[t]
    \begin{center}
    \includegraphics[width=0.95\linewidth]{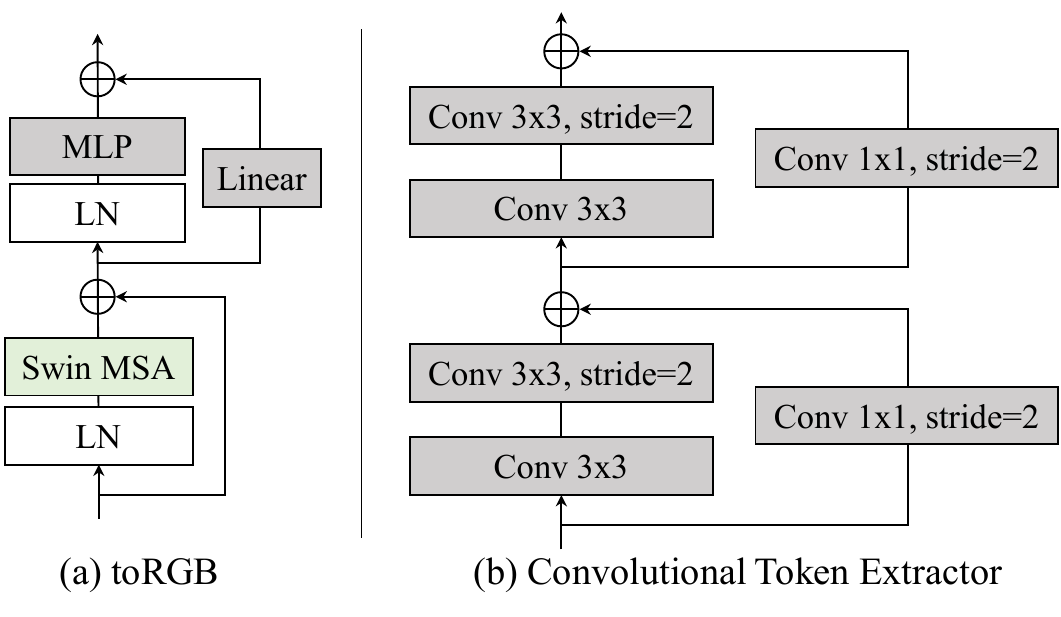}
    \end{center}
    \vspace{-5pt}
    \caption{(a) illustrates the `toRGB' layers used in our STrans-G. (b) Architecture of the convolutional residual block used to extract input tokens for discriminators.}
    \label{fig:appx-torgb}
    
\end{figure}

As shown in Fig.~2(b), in the unconditional setting, STrans-G adopts a latent code $z\sim \mathcal{N}(0, I)$ as the input, and then a linear projection layer projects it into a $(4\times 4 \times C)$ feature map.
Each spatial vector in this low-resolution feature is regarded as an input token to the following Transformer blocks.
To inject positional information~\cite{karras2021alias,xu2021positional}, we add a learnable positional encoding to the input tokens.
STrans-G gradually increases the resolution of the feature map through multiple Transformer blocks and upsampling operations. 
Except for the $64\times 64$ stage, we choose the bilinear upsampling operation in each stage. In the $64^2$ scale, we adopt the pixel shuffle~\cite{shi2016real} operator to upsample features and reduce the channel dimension.
The last `toRGB' block generates image tensors with three channels.
As shown in Fig.~\ref{fig:appx-torgb}(a), we let the MLP module output a three-channel feature and add a linear layer on the skip connection to match the channel dimension.

In STrans-G, we initialize all of the weights with a truncated normal distribution~\cite{hanin2018start}, following the original setting used in \cite{liu2021swin}.
The number of attention heads and the window size in Swin blocks are both set to 4 by default.
Different from the current vision Transformer, STrans-G only expands the channel dimensions in MLP blocks by 2 times.
Following the architectural design in StyleGAN, we adopt two Transformer blocks inside each stage in both generators and discriminators.

\noindent
\textbf{Window Size.}
In our experiments, we notice that the window size in the Swin architecture can influence the generation quality.
As shown in Tab.~\ref{tab:appx-ws}, too small window size causes a significant drop in performance. A large window size like 16 also increases the FID while introducing heavier computational costs. Consequently, we choose 4 as the default window size.
We also note that a special tuning in the architecture can lead to better performance in the FFHQ dataset, as shown in Tab.~\ref{tab:appx-ws}.
In detail, we adopt a window size of 8 in the stages from $16\times 16$ to $64\times 64$ and a window size of 4 in the stages from $128\times 128$ to $256\times 256$.
The motivation behind this design is that the layers in lower resolutions need a larger window size to learn global structures.
In the higher resolutions, it is more important to keep the locality with a small window size.

\begin{table}[t]
    \footnotesize%
    \centering
    \caption{
    We report the FID from STrans-G with different window sizes in the attention module.
    }
    \label{tab:appx-ws}
    \begin{tabular}{|c|c|c|}
        \hline
        \textbf{Window Size} & \textsc{CelebA} & FFHQ \\ \hline \hline
            2       &   7.92     &  11.35 \\
            4       &   \textbf{2.03}     & 4.98  \\
            8      &   2.85 &  5.23 \\
            16      &  3.51  &  6.14 \\
     Tuning       &   --   &  \textbf{4.84}  \\ \hline
        \end{tabular}
\end{table}

\noindent
\textbf{Model Size.}
Table \ref{tab:appx-model-size} presents a comparision of the model size among different methods.
From the experiments in StyleGAN2 and BigGAN, it is clear that more parameters will bring improvement in the synthesized quality.
It is noteworthy that our STrans-G only contains \textbf{20M} parameters in FFHQ256 dataset and \textbf{31M} parameters in \textsc{ImageNet} dataset, while the state-of-the-art generators in these two datasets adopt 46M and 70M parameters, respectively. 
%
%
Thus, the gap between STrans-G and SOTA methods partially comes from the model size. We will further try a larger model once the computational resource meets the requirements. 

\noindent
\textbf{Throughput.}
Table \ref{tab:appx-model-size} also represents the comparison in throughout. Since adopting the computation-unfriendly attention layers, STrans-G performs a slower inference speed than the traditional CNN-based generators. However, it is noteworthy that our STrans-G is trained and tested in FP32. Switching to FP16 will further increase the throughput.

\begin{table}[t]
    \footnotesize%
    \centering
    \caption{
        We compare different generators in terms of the number of parameters and also report the throughput mesured in a single Tesla V100 GPU. `ch' indicates the basic channal number introduced in StyleGAN2~\cite{karras2020analyzing} and BigGAN~\cite{brock2018large}. `2$\times$ch' represents the commonly used setting for StyleGAN2, while `1$\times$ch' is the official baseline with `channel multiplier=1' in StyleGAN2~\cite{karras2020analyzing}. We note that the official implementation of StyleGAN2 adopts the FP16 technique to accelerate the inference and training speed, while we only use the FP32 to train our STransGAN in this study. $\dag$ indicates that the number is reported in their paper.}
    \label{tab:appx-model-size}
    \begin{adjustbox}{width=\linewidth,center}
    \begin{tabular}{|c|c|c|c|c|c|}
        \hline
        \multirow{2}{*}{\textbf{Method}} & \multirow{2}{*}{Dataset} & \multirow{2}{*}{\begin{tabular}[c]{@{}c@{}}Parameters\\ (million)\end{tabular}}  & \multirow{2}{*}{FID}  & \multirow{2}{*}{\begin{tabular}[c]{@{}c@{}}Throughput\\ (imgs/sec)\end{tabular}} \\
                                &       &                                                                          &                                                                                                    &   \\ \hline\hline
            StyleGAN2, 1$\times$ch&  \multirow{4}{*}{FFHQ256}&             24.77                                                                             &    4.27  & 99.7 \\
            StyleGAN2, 2$\times$ch&  &             30.03                                                                             &    3.62 & 90.3 \\
           HiT$^\dag$~\cite{zhao2021improved} &       &       46.22                                                                          &  2.95      &   52.1  \\
           our STrans-G        & &    19.91                                                                                                                                                        &  4.84       &  63.9  \\ \hline \hline
             BigGAN ch=64     &\multirow{3}{*}{\textsc{ImageNet}128}&        31.98                                                                                                                                   &    10.48    &   56.79   \\ 
             BigGAN ch=96     &                                                                                 &   70.43                                                                         &      8.51     & 50.28  \\ 
              our STrans-G & & 30.95                                                                           &   \imgnetfid  &  41.38 \\ \hline
        \end{tabular}
    \end{adjustbox}
\end{table}

\newcommand{\normtable}{
    \begin{tabular}{|c|c|c|c|}
        \hline
        \textbf{Norm} & Dataset           & FID & IS \\ \hline \hline
        IN   & \multirow{2}{*}{CIFAR10} & 2.77  &  11.62  \\ \cline{1-1}
        BN   &                   &   5.61  &  9.55  \\ \hline \hline
        IN   & \multirow{2}{*}{\textsc{ImageNet}128} &  116.26   &  7.50  \\ \cline{1-1}
        BN   &                &   \imgnetfid    & \imgnetis    \\ \hline
        \end{tabular}
}

\begin{figure*}[t]
    \footnotesize%
    \newcommand{\h}{0.4\linewidth}%
    \newcommand{\hh}{0.295\linewidth}%
    \newcommand{\hhh}{0.01\linewidth}%
    \newcommand{\vvnew}{32.9mm}%
    \newcommand{\vvv}{0.99}%
    \subtabletopmod{\h}{\vvnew}{\vvv}{\normtable}{10pt}\hfill%
    \parbox[b]{\hh}{
    \includegraphics[width=\linewidth]{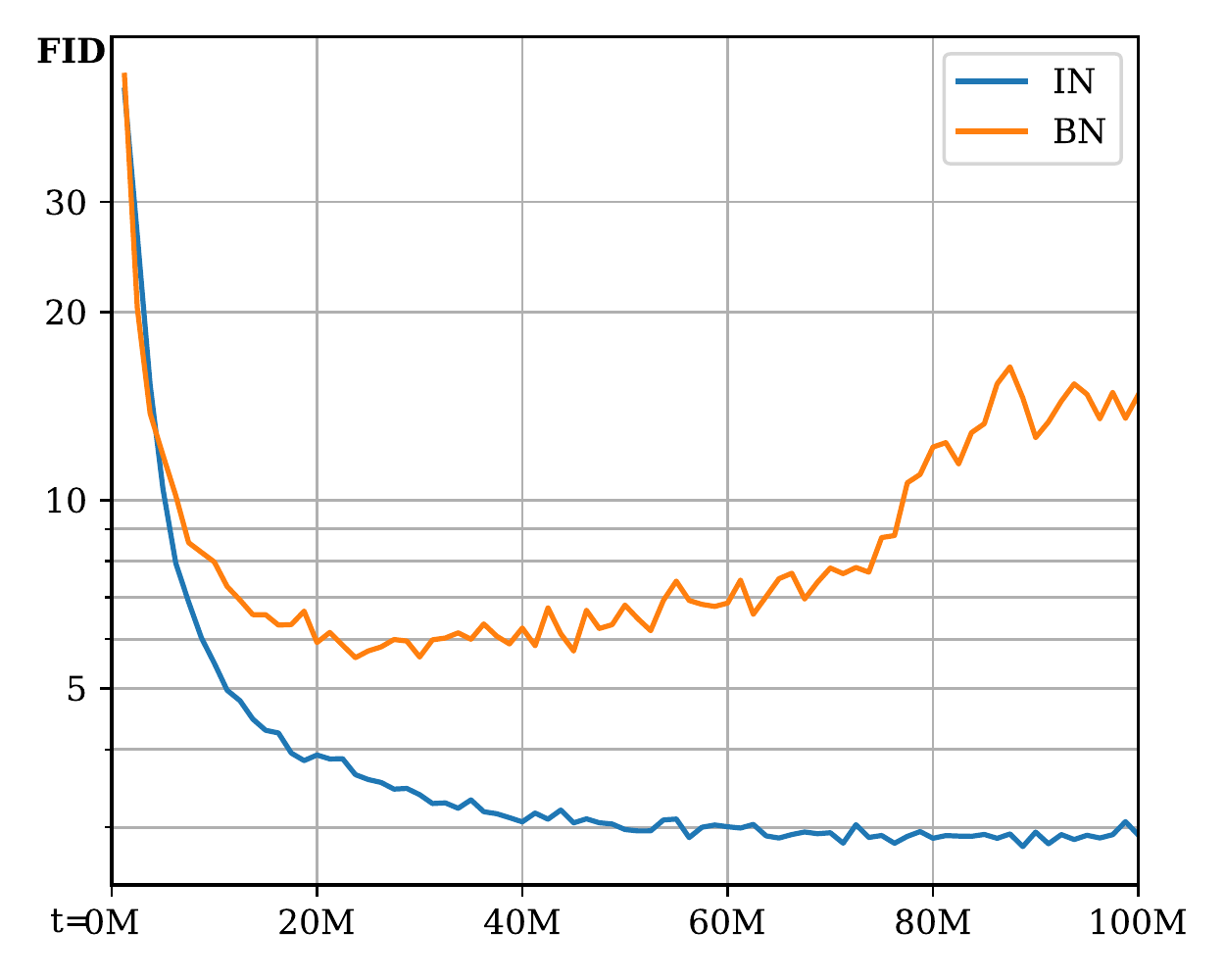}}%
    \hspace{\hhh}%
    \parbox[b]{0.295\linewidth}{
    \includegraphics[width=\linewidth]{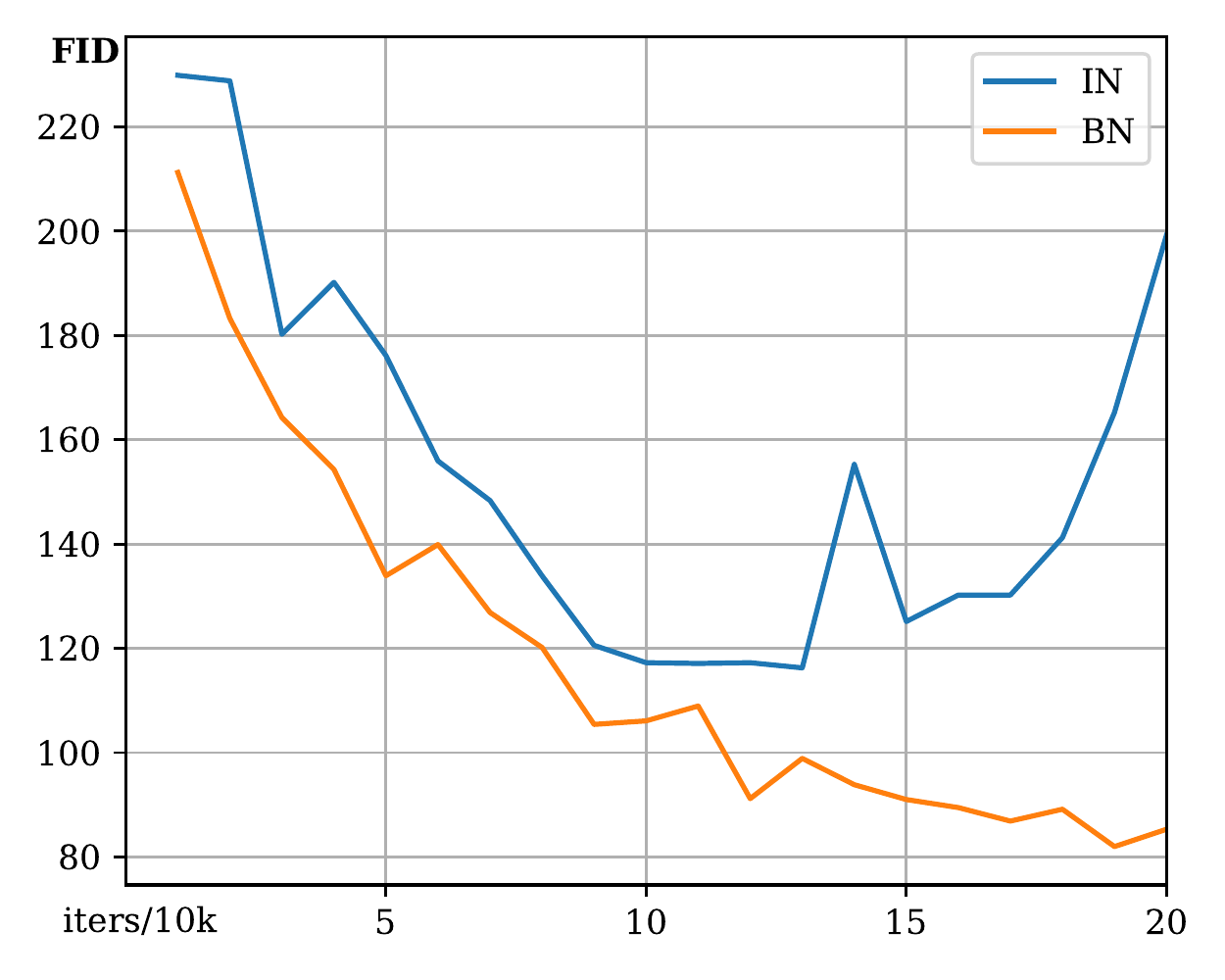}} \\
    \makebox[\h][c]{(a) Normalization in AdaNorm-T}\hfill%
    \makebox[0.3\linewidth][c]{(b) Training in CIFAR10}\hfill%
    \makebox[0.3\linewidth][c]{(c) Training in \textsc{ImageNet}}%
    \caption{
    (a) We show how different normalization layers in AdaNorm-T influence the quality of conditional generation in different datasets.
    To further show the influence in the training process, (b) plots the FID curve during training in CIFAR10.
    In (c), we only plot the training process at the early stage to show the differences between AdaIN-T and AdaBN-T in \textsc{ImageNet}.
    }
    \label{fig:appx-adanorm}
\end{figure*}

\noindent
\textbf{Normalization in AdaNorm-T.}
The choice of normalization layers in AdaNorm-T is important and related to the dataset.
In this work, we adopt instannce normalization (AdaIN-T) in CIFAR10 and batch normalization (AdaBN-T) in \textsc{ImageNet}.
Figure \ref{fig:appx-adanorm} presents how the normalization layers influence the final performance and the training process in the conditional generator, STrans-G.
In CIFAR10 with limited data, if we adopt AdaBN-T in STrans-G, the FID starts to deteriorate at the early stage.
In contrast, in \textsc{ImageNet} with large-scale data, AdaIN-T always leads to an unstable training and model divergence.
We think that it is because batch normalization layers need amounts of data and large batch sizes to model accurate statistics in the training process.
Instance normalization adopting each sample as input cannot benefit from the large batch size in \textsc{ImageNet}. 
The quantitative results in Fig.~\ref{fig:appx-adanorm}(a) also demonstrate that the choice of normalization layers highly relies on the property of the datasets.

\subsection{STrans-D}
\label{sec:appx-strans-d}



\noindent
\textbf{Hybrid Design.}
We directly adopt the residual downsampling blocks in the StyelGAN2 discriminator as our feature extractor generating input tokens.
As shown in the  Fig.~\ref{fig:appx-torgb}(b), each convolutional block downsamples the input features by $2\times$ times. We apply two successive convolutional blocks outputting the input tokens in a $4\times$ downsampled scale.
By stacking Swin Transformer blocks with bilinear downsampling operation, the high-resolution input tokens are gradually downsampled to ($4\times 4 \times C$) feature.
Note that we discard the original patch merging module in \cite{liu2021swin} to avoid a large channel dimension. 
Containing a convolutional layer and several linear layers, the critic head in Fig.~4(a) outputs the final prediction value.

\noindent
\textbf{Equalized Learning Rate.}
The equalized learning rate (EqLR) is originally designed for scaling the weight at runtime instead of applying careful weight initialization, like $\mathcal{N}(0, 0.02)$.
In StyleGAN, \cite{karras2019style} extend EqLR to the style mapping network to adjust the learning rate separately.
In STrans-D, we observe that the Transformer blocks need a small learning rate to keep a stable training procedure.
However, a slow learning speed in the discriminator always brings slow convergence speed in the generator and unsatisfactory generation qualities, which is also a well-known practice~\cite{heusel2017gans} in current literature. 
Thus, to keep the whole discriminator trained with a large learning rate, we adopt EqLR in the sub-layers of the attention and pointwise fully connected operations.
In precise, we rescale the learnable weights $w_i$ by multiplying a small scaler $c_{lr}$ in the forward pass:
\begin{equation}
    \hat{w}_i = c_{lr} \cdot w_i. 
\end{equation}
Then, the rescaled $\hat{w}_i$ will be used in the subsequent operations like the linear projection.

In our experiments, without EqLR, the config-a and config-b in Fig~7(a) are trained with a small learning rate of 0.0001. Otherwise, the adversarial training will diverge in the early stages. After adopting EqLR, we apply a normal learning rate of 0.002 in the whole discriminator.

\noindent
\textbf{Initialization.}
In STrans-D, we initialize the weights in the convolutional feature extractor and the critic head with standard $\mathcal{N}(0, I)$ distribution. For the Swin Transformer blocks, we still use the truncated normalization following the original implementation. 
However, we slightly increase the value range of the initialization strategy, because the equalized learning rate module in these blocks contains a learning rate scaler of 0.1.
In detail, we select $\mathcal{N}(0, 0.2)$ as the default distribution and truncated the value in the range of $[-10, 10]$.

\begin{table}
    \centering
    \caption{We present the influence of differentiable data augmentation on the final FID. With STrans-G as the generator, we also compare the behavior of StyleGAN2 discriminator (`CNN') and our STrans-D (`Trans'). }
    \label{tab:appx-ada}
    \begin{adjustbox}{width=0.25\textwidth,center}
    \begin{tabular}{|l|c|cc|}
        \hline
        \multirow{2}{*}{\textbf{Dataset}} & \multirow{2}{*}{Disc} & \multicolumn{2}{c|}{ADA} \\
                                 &                       & w/o         & w/         \\ \hline \hline
                                 \multirow{2}{*}{\textsc{CelebA}}                & CNN             &  2.41     &     2.03     \\
                                 & Trans             & 3.32&3.16\\ \hline
        \multirow{2}{*}{FFHQ}    & CNN             &    5.24    &  4.84      \\
                                 & Trans             &   6.22   & 5.98  \\ \hline
        CIFAR10                  & CNN             &     4.07       &  2.77           \\ \hline
    \end{tabular}  
    \end{adjustbox} 
    \vspace{-10pt}
\end{table}
\begin{figure*}
    \centering
    \includegraphics[width=\linewidth]{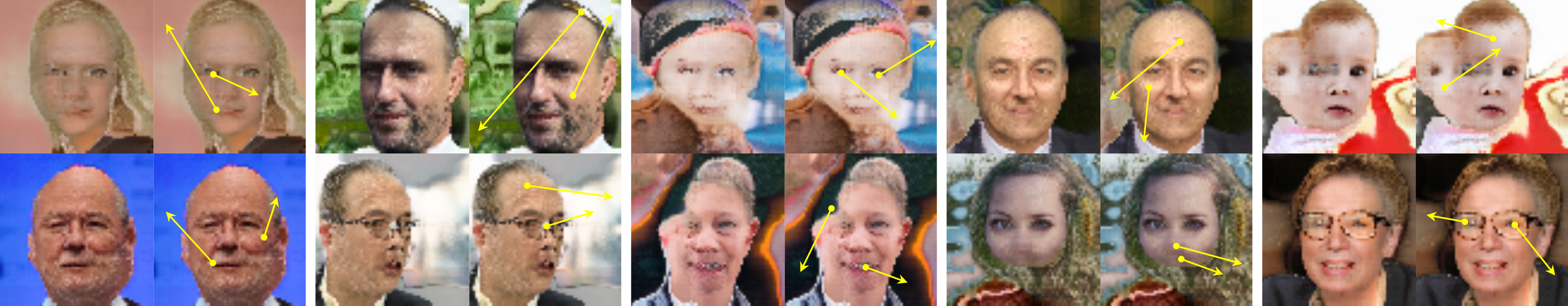}
    \caption{Additional results for the generated $64\times 64$ FFHQ samples from Trans-G. Meanwhile, we analyze the attention behavior in each sample. Given a query pixel at the starting point, the end point of the yellow arrow in this figure represents the pixel with the highest attention score in the last attention layer.}
    \vspace{-10pt}
    \label{fig:supp-transg}
\end{figure*}

\subsection{Effects of Differentiable Data Augmentation}
\label{sec:appx-ada}
Recent concurrent works~\cite{jiang2021transgan,lee2021vitgan,zhao2021improved} claim that differentiable data augmentation~\cite{karras2020training,zhao2020differentiable,zhao2020improved} plays a more important role in Transformer-based generative models than in CNN-based ones.
Different from these findings, as shown in Tab.~\ref{tab:appx-ada}, STransGAN with Swin architecture can achieve competitive performance without differentiable data augmentation in regular datasets, \ie, \textsc{CelebA}, and FFHQ.
Even if in CIFAR10 where the training data is highly limited, removing differentiable data augmentation will not cause any serious drop of FID.
This implies that our STransGAN can learn the complex image distribution in a data-efficient way.

\subsection{Evaluation}
For FFHQ and \textsc{CelebA} dataset, we calculate the Fr\'{e}chet inception distance (FID) between 50k generated samples and 50k real samples randomly selected from the training set.
Following the implementation in \cite{brock2018large}, we compute FID with 50k generated samples and the full training set in CIFAR10 and \textsc{ImageNet}.
We also sample 50k images to evaluate IS metric.
Note that we report the IS score with the checkpoint obtaining the best FID in all of our experiments.

\setcounter{subsection}{0}
\section{Additional Results}
\label{app:res}
In Fig.~\ref{fig:supp-transg}, we show more failure cases from Trans-G and analyze their attention maps in the same way as Fig.~3(c). These additional results show convincing evidence that global attention is not as accurate as we expect. When synthesizing high-resolution features, such an unhealthy behavior can easily mislead the generative moels causing unsatisfactory artifacts. We notice that some generated samples in Fig.~3(a) and Fig.~3(b) contains alising artifacts. This may caused by the large downsampling ratio, $64/1024=0.0625$. In data pre-processing, we have followed the suggestion used in \cite{karras2017progressive} where they adopte the anti-aliasing interpolation method from PIL.

\begin{figure*}[t]
    \begin{center}
    \includegraphics[width=0.95\linewidth]{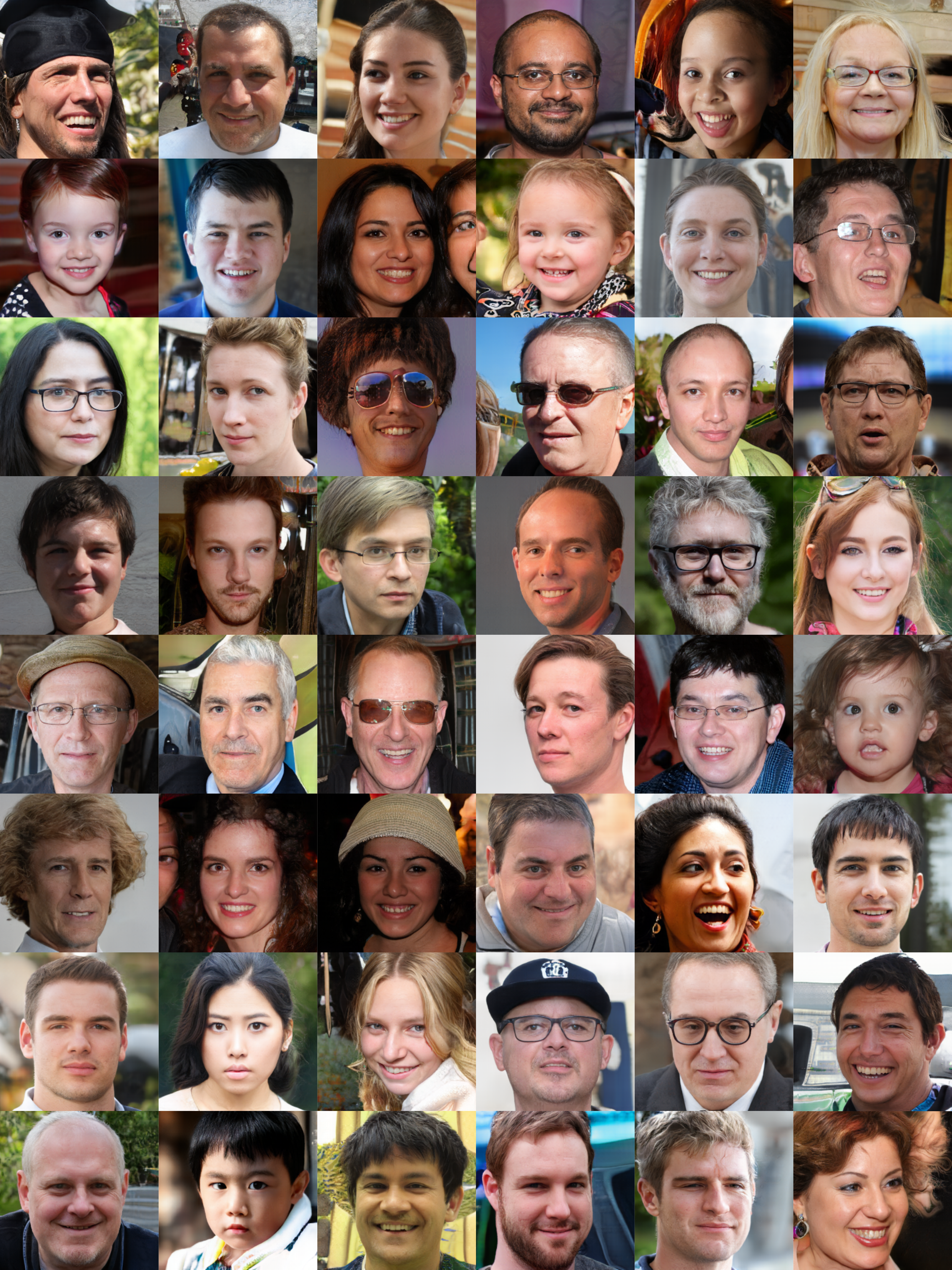}
    \end{center}
    \vspace{-5pt}
    \caption{Unconditional FFHQ $256^2$ samples from our STrans-G.}
    \label{fig:appx-ffhq}
\end{figure*}

\begin{figure*}[t]
    \begin{center}
    \includegraphics[width=\linewidth]{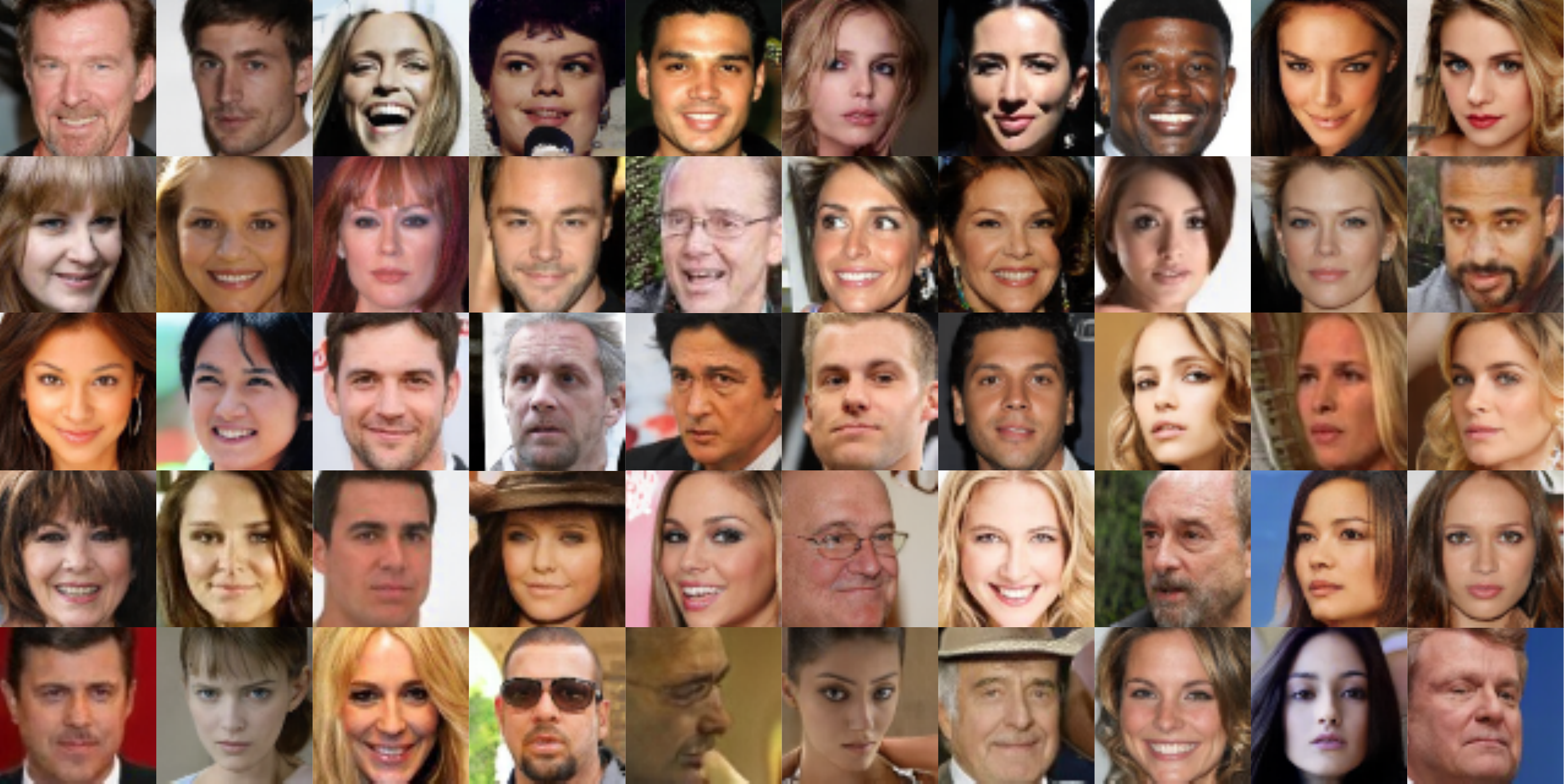}
    \end{center}
    \vspace{-5pt}
    \caption{Unconditional \textsc{CelebA} $64^2$ samples from our STrans-G.}
    \label{fig:appx-celeba}
\end{figure*}

\begin{figure*}[t]
    \begin{center}
    \includegraphics[width=\linewidth]{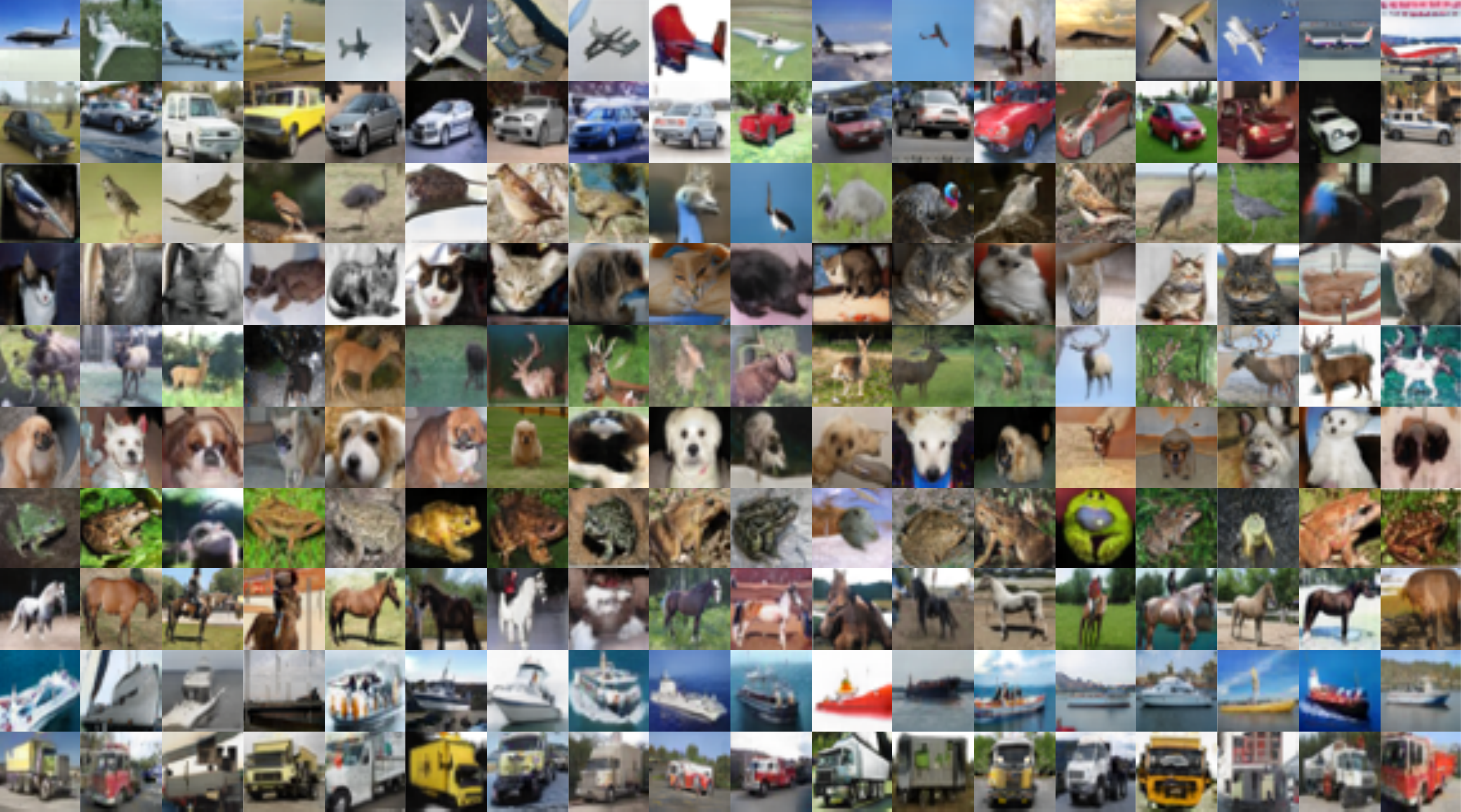}
    \end{center}
    \vspace{-5pt}
    \caption{Conditional CIFAR10 $32^2$ samples from our STrans-G with AdaIN-T. Each row presents the samples from one category in CIFAR10.}
    \label{fig:appx-cifar10}
\end{figure*}

\begin{figure*}[t]
    \begin{center}
    \includegraphics[width=0.95\linewidth]{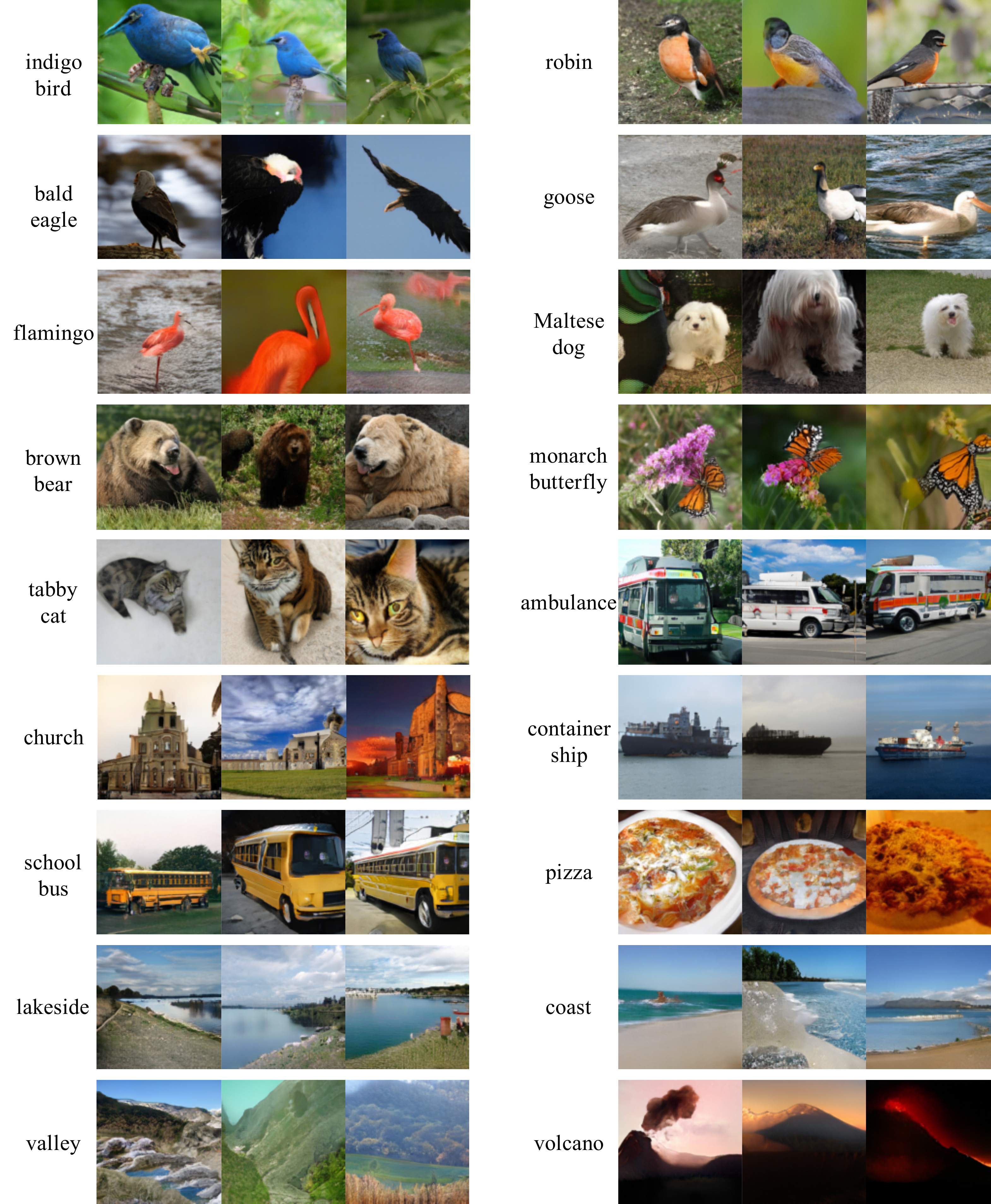}
    \end{center}
    \vspace{-5pt}
    \caption{Conditional \textsc{ImageNet} $128^2$ samples from our STrans-G with AdaBN-T. We adopt a truncation threshhold of 0.5 when performing random sampling. }
    \label{fig:appx-imgnet}
\end{figure*}

\begin{figure*}[t]
    \begin{center}
    \includegraphics[width=0.73\linewidth]{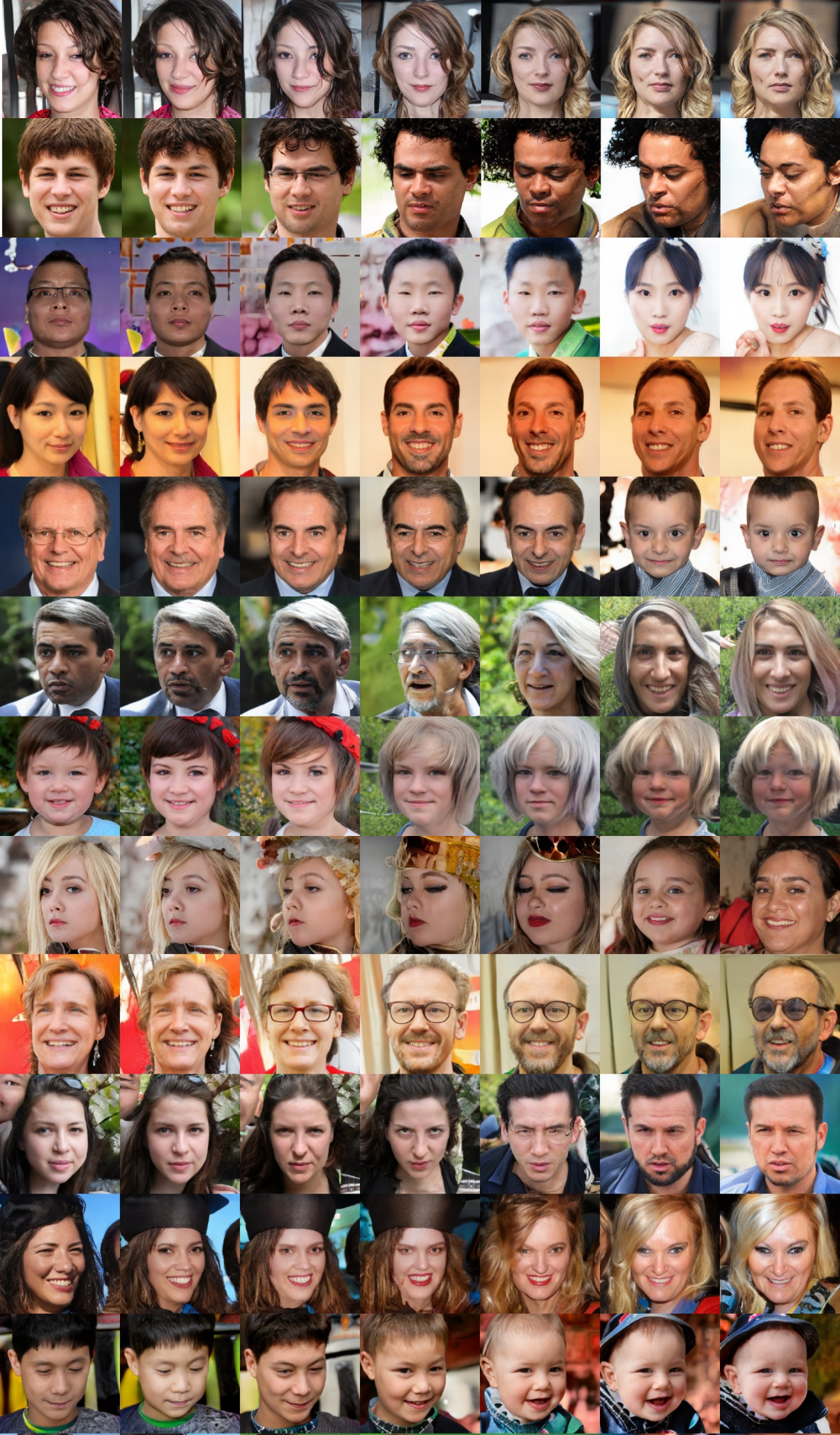}
    \end{center}
    \vspace{-15pt}
    \caption{Examples for latent space interpolation in FFHQ $256^2$.}
    \label{fig:appx-interp-ffhq}
\end{figure*}

In Fig.~\ref{fig:appx-ffhq}, Fig.~\ref{fig:appx-celeba}, Fig.~\ref{fig:appx-cifar10}, and Fig.~\ref{fig:appx-imgnet}, we present additional samples from STrans-G in FFHQ, \textsc{CelebA}, CIFAR10, and \textsc{ImageNet} respectively.
We only adopt the truncation tricks for random sampling in \textsc{ImageNet} dataset.
As shown in the visualization results, the local window attention mechanism in the Swin architecture will not bring boundary artifacts in the synthesized images.
Meanwhile, the visually pleasant results indicate that STrans-G provides a promising direction of generating high-quality images with pure Transformer models.
Figure \ref{fig:appx-interp-ffhq} presents the results for the interpolation of the latent space from STrans-G trained in FFHQ $256^2$.

\section{Ethics and Reproducibility Statement}
%
\noindent\textbf{Ethics Statement.}
This work improves the image generation quality, which can facilitate the movie industry and virtual reality.
However, there could be potentially inappropriate applications of this technique, lying in various forms of disinformation, \eg, synthesizing fake portraits in different scenes~\cite{perov2020deepfacelab}. Our work may unintentionally make these applications more convincing or deceiving.
Viable solutions to prevent misusage and negative effects include developing robust methods to detect fake content generated by deep models~\cite{tolosana2020deepfakes} and adopting model watermarking~\cite{yu2020artificial}.

\noindent\textbf{Reproducibility Statement.}
We have provided the necessary implementation details in Sec.~3, Sec.~4, and Appendix \ref{app:impl}. In addition, we also include more results with different settings of STransGAN in Appendix \ref{app:impl}. Our training pipeline follows standard training configurations in current literature and does not include extra tricks. The models and codes will be publicly available.

\end{document}